\documentclass{article}
\usepackage{ijcai20}

\usepackage{natbib}
\usepackage{makecell}
\usepackage{bm}
\usepackage{cite}
\usepackage{times}
\usepackage{soul}
\usepackage{url}
\usepackage[hidelinks]{hyperref}
\usepackage[utf8]{inputenc}
\usepackage[small]{caption}
\usepackage{graphicx}
\usepackage{amsmath}
\usepackage{amsthm}
\usepackage{booktabs}
\usepackage{algorithm}
\usepackage{algorithmic}
\usepackage{amsfonts}
\usepackage{enumitem}
\usepackage{amssymb}
\usepackage{latexsym}
\usepackage{autobreak}
\usepackage[labelformat=simple]{subcaption}
\usepackage{bbding}
\usepackage{pifont}
\usepackage{placeins}
\usepackage{footmisc}
\renewcommand{\thefootnote}{*}
\graphicspath{{./images/}}

\usepackage{stfloats}
\usepackage{multirow}
\usepackage{mathrsfs}

\title{Improving the Transferability of Time Series Forecasting with Decomposition Adaptation}

\author{
Yan Gao $^{1}$\footnotemark[2]\and
Yan Wang $^{2}$\footnotemark[2]\And
Qiang Wang $^{1}$\footnotemark[1]
\affiliations
$^1$College of Computer Science and Technology, Zhejiang university, Hangzhou, China\\
$^2$School of Artificial Intelligence, Nanjing University, Nanjing, China\\
\emails
\ 22021022@zju.edu.cn, yanwang@smail.nju.edu.cn,wangqiang@zju.edu.cn
}

\begin{document}
\setcounter{tocdepth}{3}
\setcounter{secnumdepth}{3}

\maketitle

\begin{abstract}
\label{abstract}
Due to effective pattern mining and feature representation, neural forecasting models based on deep learning have achieved great progress. 
The premise of effective learning is to collect sufficient data. However, in time series forecasting, it is difficult to obtain enough data, which limits the performance of neural forecasting models. 
To alleviate the data scarcity limitation, we design Sequence Decomposition Adaptation Network (SeDAN) which is a novel transfer architecture to improve forecasting performance on the target domain by aligning transferable knowledge from cross-domain datasets. 
Rethinking the transferability of features in time series data, we propose Implicit Contrastive Decomposition to decompose the original features into components including seasonal and trend features, which are easier to transfer.
Then we design the corresponding adaptation methods for decomposed features in different domains.
Specifically, for seasonal features, we perform joint distribution adaptation and for trend features, we design an Optimal Local Adaptation. 
We conduct extensive experiments on five benchmark datasets for multivariate time series forecasting. 
The results demonstrate the effectiveness of our SeDAN. It can provide more efficient and stable knowledge transfer.
\end{abstract}

\renewcommand{\thefootnote}{\fnsymbol{footnote}}
\footnotetext[2]{Y. Gao and Y. Wang contributed equally to this work.}
\footnotetext[1]{Corresponding author.}

\section{Introduction}
\label{sec:intro}

Time series forecasting is a crucial problem within machine learning. Based on the historical series, we mine the potential correlations and train models to capture the complex temporal dynamics. The related algorithms have been widely used in  trafﬁc forecasting \citep{1-traffic-DBLP:conf/iclr/LiYS018,2-traffic-DBLP:conf/ijcai/YuYZ18,3-traffic-DBLP:conf/nips/0001YL0020}, financial transactions \citep{6-financial-DBLP:conf/nips/MeiE17,7-financial-DBLP:journals/asc/SezerGO20}, and demand decision making \citep{8-retailers-DBLP:journals/corr/FlunkertSG17,9-retailers-DBLP:conf/iclr/OreshkinCCB20}. Early models are based on classical statistical methods, such as ARIMA and Linear State Space Models. These models often have a complete theoretical basis, but remain limited in the expressivity. Benefiting from the powerful expressivity of deep learning, neural forecasting models represented by Recurrent Neural Networks (RNN) \citep{10-RNN-DBLP:journals/neco/HochreiterS97,11-RNN-DBLP:journals/corr/ChungGCB14} and Temporal Convolutional Networks (TCN) \citep{12-TCN-DBLP:conf/ssw/OordDZSVGKSK16,13-TCN-DBLP:journals/corr/abs-1803-01271} have achieved impressive performance on sequence modeling.
The neural forecasting models automatically extract the high-order feature representations, but need a suﬃcient amount of data. The Transformer architecture \citep{14-transformer-DBLP:conf/nips/VaswaniSPUJGKP17} makes remarkable breakthroughs in Natural Language Processing, Computer Vision and other fields. Recently, there has been some works applying the Transformer-based models for time series forecasting. In order to enhance flexible expressivity, the Transformer-based models discard more inductive biases than CNN and RNN, thus encouraging more reliance upon the large datasets. However, unlike in other fields, it is more challenging to collect enough data for training in time series forecasting, which severely limits the performance of the Transformer-based models.

An effective way to deal with the above problems is to explore the transferability of knowledge learned by deep learning models. By learning general knowledge in different datasets, the existing data can be trained more efficiently. Some transfer learning models, represented by Deep Domain Adaptation, measure the similarity between different domains in the feature space and align their feature distributions to alleviate the distribution shift. 
We use transfer learning to reduce the impact of data scarcity on neural forecasting models. When designing transfer learning models, it is necessary to consider three basic questions \citep{17-survey-DBLP:journals/tkde/PanY10}: ``What" kind of knowledge can be transferred as general knowledge across domains, ``How" to develop a learning algorithm to transfer these knowledge and ``When" to transfer to avoid negative transfer. This paper mainly explores the transferability in time series forecasting from the perspective of ``What" and ``How".

For time series forecasting , there are two main problems in applying existing transfer learning methods directly. Firstly, we cannot obtain the datasets of sufficient size as source domain data. When the source domain data is sufficient, the basic methods such as pre-training and fine-tuning paradigm can help the model achieve satisfactory performance on downstream tasks, which is verified in CV and NLP \citep{18-bert-DBLP:conf/naacl/DevlinCLT19,19-GPT3-DBLP:conf/nips/BrownMRSKDNSSAA20,20-MoCo-DBLP:conf/cvpr/He0WXG20}. However, in time series forecasting, there is a lack of large datasets for pre-training such as ImageNet ILSVRC \citep{21-Imagenet-DBLP:journals/ijcv/RussakovskyDSKS15} and MS COCO \citep{22-COCO-DBLP:conf/eccv/LinMBHPRDZ14}. Therefore, we need a more refined adaptation to transfer knowledge from small-scale pre-training datasets to target domain. Secondly, in the process of transferring knowledge across different domains, existing domain adaptation models aim to transfer the marginal distribution and the conditional distribution at the same time. This assumption is reasonable in general classification problems, where the label space of the source and target domains are the same. However, in our task, the conditional distribution is reflected in the temporal dynamics of the latent state, which is not necessarily the same in different domains. We argue that a more stable transfer model can be obtained by excluding the hard-to-transfer knowledge related to the conditional distribution. A few recent works \citep{23-TS+fransfer-AWS-work-DBLP:journals/corr/abs-2102-06828,24-TS+fransfer-DBLP:journals/pr/YeD21} have explored the transferability in the task. However, these works do not consider the problem that the conditional distribution is difficult to transfer across domains directly.

\begin{figure}[h]
\centering
\includegraphics[width=\linewidth]{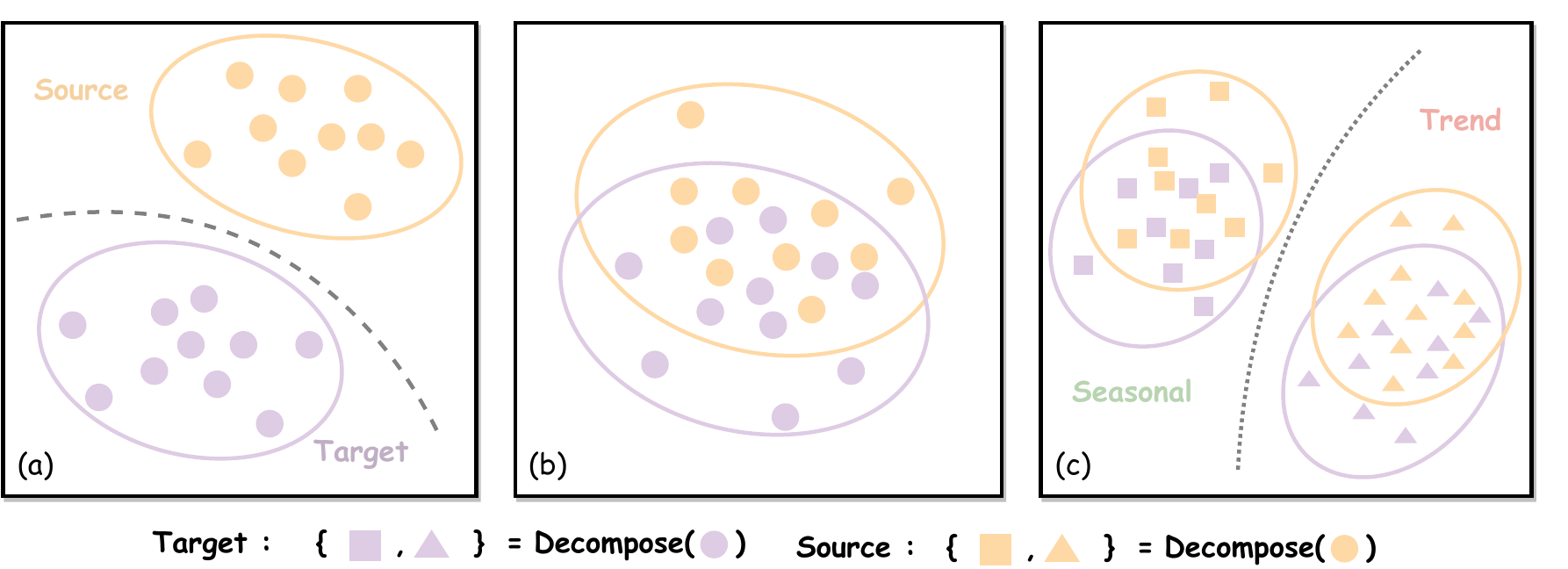}\par\caption{Comparison between previous adaptation models and our decomposition adaptation in the feature space. (a) The entangled features of source and target domain before adaptation. (b) Previous adaptation models align the entangled features directly. (c) Our method performs corresponding adaptation on the decomposed features.}
\label{fig:fig-1}
\end{figure}

Based on the above motivations, we propose a Sequence Decomposition Adaptation Network.
The core idea is that we decompose the entangled features, and then design adaptation methods respectively for the decomposed components.
Specifically, inspired by contrastive learning \citep{27-MOCO-DBLP:conf/cvpr/He0WXG20}, we design Implicit Contrastive Decomposition (ICD), which decomposes the encoded features into seasonal features component and trend features component for further adaptation. During the decomposition process, the transferable part and non-transferable part of the conditional distribution are also separated. The seasonal features contain marginal distribution and transferable conditional distribution. For seasonal features, we perform joint probability adaptation. In the meanwhile, the trend features contain marginal distribution and non-transferable conditional distribution. Therefore, we design Optimal Local Adaptation (OLA) for the trend feature, which formalizes the adaptation problem as a global optimal matching problem, so as to realize that only the temporal-invariant distribution (marginal distribution) will be transferred during adaptation.
The decomposition adaptation in the feature space is shown in Figure~\ref{fig:fig-1}.
The SeDAN follows the encoder-decoder structure and utilizes the vanilla Transformer as the base model. The main contributions of this work are as follows:

\begin{enumerate}
\item For cross-domain transfer in time series forecasting, we propose a SeDAN model to learn general knowledge including marginal distribution and the transferable part of conditional distribution of features. We adapt the decomposed features respectively with joint probability adaptation and a novel Optimal Local Adaptation to avoid learning non-transferable knowledge.
\item To decompose features for further adaptation, we design feature-level Implicit Contrastive Decomposition, utilizing the idea of contrastive learning to generate seasonal features component and trend features component.
\item We conduct extensive experiments to demonstrate the performance of the SeDAN on five multivariate time series forecasting datasets. The experimental results verify the effectiveness and stability of the SeDAN. We also use visualization methods to show the significance of decomposition adaptation in feature space.
\end{enumerate}
\section{Related Work}
\label{sec:related_work}

\begin{figure*}[htp]
\centering
\includegraphics[width=0.9\textwidth]{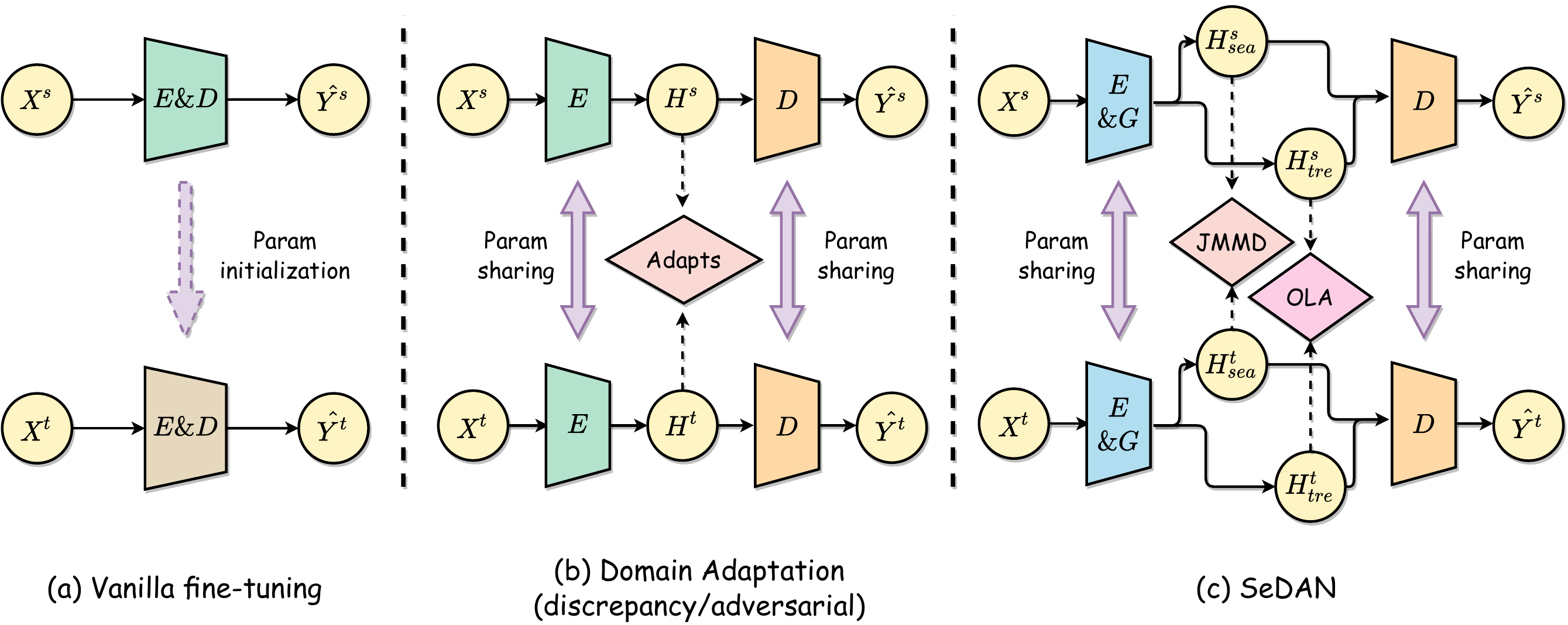}\par
\caption{Comparison on structures between the existing transfer learning models and our proposed SeDAN. (a) Vanilla fine-tuning: trains a model on target domain but initialising it with pre-trained weights obtained from training it on the source domain. (b) Domain Adaptation: designs the discrepancy-based or adversarial-based adaptation models to reduce the distance of distributions across domains at the feature-level. (c) Our SeDAN: performs the corresponding adpation methods (JMMD for seasonal features and OLA for trend features) to adapt the decomposed features across domains.}
\label{fig:fig-2}
\end{figure*}

\paragraph{Time Series Forecasting}
Nseural forecasting models begin to show advantages in handling data with complex patterns. There are three main types of existing neural forecasting models, which are RNN-based models \citep{10-RNN-DBLP:journals/neco/HochreiterS97,11-RNN-DBLP:journals/corr/ChungGCB14,32-RNN-LSTNet-DBLP:conf/sigir/LaiCYL18}, TCN-based models \citep{12-TCN-DBLP:conf/ssw/OordDZSVGKSK16,13-TCN-DBLP:journals/corr/abs-1803-01271,33-trellisnet-DBLP:conf/iclr/BaiKK19} and Transformer-based models \citep{15-informer-DBLP:conf/aaai/ZhouZPZLXZ21,16-TFT-DBLP:journals/corr/abs-1912-09363,34-trasnfoermer-based-DBLP:conf/nips/WuXDZ0H20,35-trasnfoermer-based-DBLP:conf/nips/LiJXZCWY19}. 
The RNN-based models often utilize gated units to alleviate the long-term dependency \citep{15-informer-DBLP:conf/aaai/ZhouZPZLXZ21} and follow the  Seq2Seq models \citep{35-trasnfoermer-based-DBLP:conf/nips/LiJXZCWY19,37-seq2seq-DBLP:journals/corr/abs-2010-07349} for multi-horizon forcasting. Another way to relieve the long-term dependency is to employ skip connections \citep{38-rnn-DBLP:conf/nips/ChangZHYGTCWHH17} or multi-scale dependency \citep{39-rnn-DBLP:conf/ijcai/ChenML21}. However, the sequential dependency issue of the RNN-based models makes it hard to parallelize. 
Through the causal convolution, TCN maintains parallelism in processing time series data, but the model scalability in handling long-term dependency still remains limited. 
Through self-attention mechanism, the Transformer-based models \citep{14-transformer-DBLP:conf/nips/VaswaniSPUJGKP17} reduce the traveling path between the current signal and the historical signal to $O(1)$. However, the high computational complexity becomes the bottleneck. To reduce the computational cost, prior works \citep{25-shift-DBLP:conf/cvpr/TorralbaE11,40-reformer-DBLP:conf/iclr/KitaevKL20} design the unstructured sparse attention by introducing local correlation assumptions. \citeauthor{15-informer-DBLP:conf/aaai/ZhouZPZLXZ21} propose Informer with a structured sparse attention from the perspective of activity differences for the distribution.

\paragraph{Transfer Learning and Domain adaptation}
Transfer learning relaxes the I.I.D. assumption between training data and test data, aiming to learn general knowledge from the source domains which are similar to the target domain. The knowledge transfer in time series forecasting is related to the Domain Adaptation (DA) \citep{41-DA-DBLP:journals/corr/TzengHZSD14}, which assumes that the marginal distribution of data between source and target domain is different. There are three types of methods to deal with DA \citep{42-DA-survey-DBLP:journals/tnn/ZhaoYZLZWKGSSK22}, including Discrepancy-based, Adversarial-based and Generative-based methods. The Discrepancy-based and Adversarial-based methods both focus on reducing the distance of feature-level distributions between two domains. The difference is that the Discrepancy-based methods use a explicit metric function to define the distance between two domains. In contrast, the Adversarial-based methods use the adversarial training to replace the metric function with a neuralized discriminator.

Recently there has been some works focusing on knowledge transfer in time series \citep{23-TS+fransfer-AWS-work-DBLP:journals/corr/abs-2102-06828,24-TS+fransfer-DBLP:journals/pr/YeD21,43-TSC-transfer-DBLP:journals/corr/abs-2109-14778,44-TSC-transfer-DBLP:journals/corr/abs-2111-14834,45-TSC-transfer-DBLP:conf/ijcai/LiuX21,46-TS-transfer-DBLP:conf/aaai/CaiC0CZYLYZ21}. \citeauthor{45-TSC-transfer-DBLP:conf/ijcai/LiuX21} propose a hybrid spectral kernel network based on spectral theorem and kernel approximation to perform alignment across domains. \citeauthor{43-TSC-transfer-DBLP:journals/corr/abs-2109-14778} utilize adversarial training to align the feature-level distribution across domains, and design a contrastive learning algorithm to leverage cross-source label information. However, these models only focus on time series classification. For the time series forecasting, \citeauthor{24-TS+fransfer-DBLP:journals/pr/YeD21} utilize the Dynamic Time Warping and Jensen-Shannon to measure the similarity between datasets and embed the transferring stage into the feature learning process. Further, \citeauthor{23-TS+fransfer-AWS-work-DBLP:journals/corr/abs-2102-06828} take advantage of attention modules and adversarial training to learn the domain-independent features. These methods address partial problems of applying transfer learning to time series forecasting, but do not consider that the stochastic transition dynamics of latent state is hard to transfer. We design a model to decompose the hard-to-transfer conditional distribution, followed by further distribution adaptation. The comparison between our proposed decomposition adaptation model and the existing transfer learning models is shown in Figure~\ref{fig:fig-2}.

\paragraph{Decomposition}
Time series data can be decomposed into a series of components \citep{47-decomposition-STL-cleveland1990stl,48-decomposition-EMD-huang1998empirical,49-decomposition-X11-shiskin1967x,50-decomposition-DBLP:journals/tnn/GodfreyG18}. Due to the reduced coupling, feature patterns are more easier to extract. Learning from different components respectively can improve the prediction performance of the model. \citeauthor{47-decomposition-STL-cleveland1990stl} design a Seasonal and Trend Decomposition Procedure based on Loess (STL), using robust locally-weighted regression as a smoothing method.  \citeauthor{48-decomposition-EMD-huang1998empirical} propose an Empirical Mode Decomposition mothod that is not based on basis functions but only relies on local characteristic time scale. The above methods all design an explicit structure to decompose sequences. \citeauthor{50-decomposition-DBLP:journals/tnn/GodfreyG18} present a neural decomposition model, which uses neural networks to conduct the decomposition of periodic and aperiodic terms. Different from our work, this model only handles univariate sequence in the time domain. We propose ICD that can handle more complex multivariate sequence in feature level.

\paragraph{Contrastive Learning}
Contrastive learning is a type of self-supervised learning \citep{51-CL-survey-DBLP:journals/corr/abs-2006-08218}, which learns similarity between samples with the same attribute. Related works have been used for many tasks of time series \citep{43-TSC-transfer-DBLP:journals/corr/abs-2109-14778,54-TS-CL-DBLP:conf/ijcai/Eldele0C000G21,55-TS-CL-DBLP:conf/iclr/TonekaboniEG21,56-TS-CL-DBLP:conf/nips/FranceschiDJ19}. \citeauthor{55-TS-CL-DBLP:conf/iclr/TonekaboniEG21} propose a Temporal Neighborhood Coding that exploits local smoothness during temporal signal generation to learn generalizable representations of windows. \citeauthor{54-TS-CL-DBLP:conf/ijcai/Eldele0C000G21} design a context-based contrast module, which maximizes the similarity between different contexts in the same sample while minimizing the similarity between the contexts in different samples to learn robust temporal representations. In this paper, our proposed model utilizes contrastive learning to decompose features instead of learning the general feature representations.

\section{Preliminary}
\label{sec:preliminary}

This paper focuses on the Transfer Learning in time series forecasting. Before introducing our method, we provide the problem deﬁnition. Single-source transfer learning is discussed in this paper. Given a set of time series data as the target domain data $D^t = \{ X^t, Y^t \}$, we have $ X^t = ( x_1^t, x_2^t, \cdots, x^t_{l_{tx}} ) \in \mathbb{R}^{l_{tx} \times d_{tx}}$ and $ Y^t = (y_1^t, y_2 ^t, \cdots, y^t_{l_{ty}}) \in \mathbb{R}^{l_{ty} \times d_{ty}} $, where $d_{tx} \geqslant 1$ and $d_{ty} \geqslant 1$ represent the dimension of time series data. In addition , we have a set of source domain data $D^s = \{ X^s, Y^s \}$, where $ X^s \in \mathbb{R}^{l_{sx} \times d_{sx}} $ and $ Y^s \in \mathbb{R}^{l_{sy} \times d_{sy}} $. The dimensions $d_{tx} $ and $d_{sx} $, $d_{ty} $ and $d_{sy} $ are not necessarily equal. Both of $D^s$ and $D^t$ are time series data, and we assume that they share the same feature space, which can be expressed as $\mathcal{X}^{s}=\mathcal{X}^ {t}$, but have different marginal distributions $P_{s}\left(X^{s}\right) \neq P_{t}\left(X^{t}\right)$ because of the Covariate Shift [57]. Besides, due to the different stochastic transition dynamics of latent state, we consider that the conditional distributions are also different: $P_s \left(Y^s|X^s \right) \neq P_t \left(Y^t|X^ t \right)$ , which leading to the Dataset Shift [58] in the task. Our goal is to exploit both $D_t$ and $D_s$ to learn forecasting model for the target domain, mitigating the impact of Dataset Shift.
\section{Methodology}
\label{sec:method}

In this section, we propose a Sequence Decomposition Adaptation Network. We introduce the overall structure of SeDAN in subsection~\ref{ssec:4-1}, then, we describe the key modules: the Seasonal-Trend Decomposition module for feature sequences (in subsection~\ref{ssec:4-2}), and the Decomposition Adaption module for the decomposition components (in subsection~\ref{ssec:4-3}). 

\subsection{Overall Framework}
\label{ssec:4-1}

\begin{figure*}[htp]
\centering
\includegraphics[width=\textwidth]{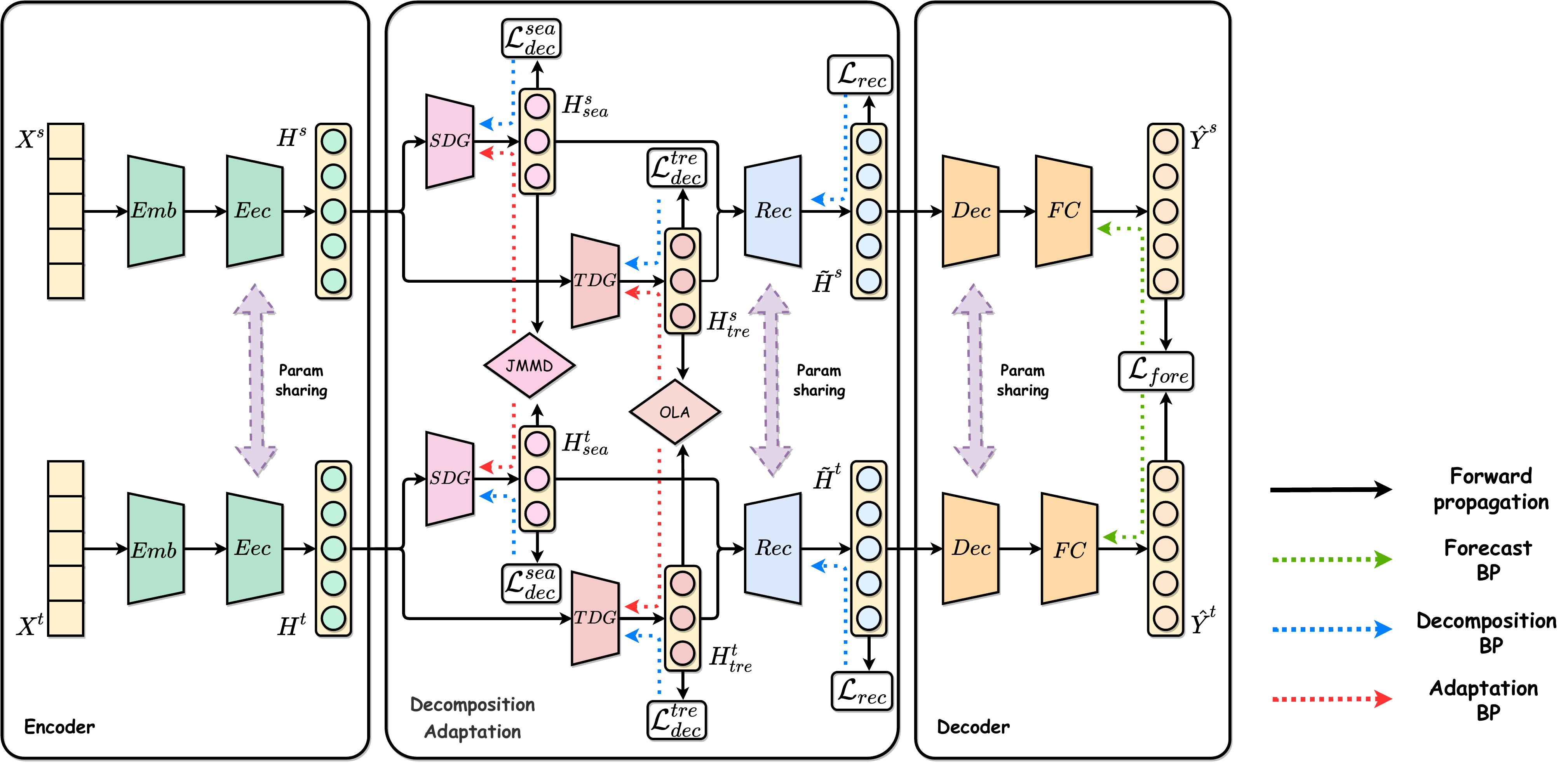}\par
\caption{Overall framework of our SeDAN model. We design a Decomposition Adaptation method for knowledge transfer in time series forecasting, containing three parts: the basic Attention-based prediction module (including encoder and decoder), the Decomposition module and the Adaptation module. The dashed arrows describe the training process of the network.}
\label{fig:fig-3}
\end{figure*}

As shown in Figure~\ref{fig:fig-3}, the SeDAN model consists of three main components, including the basic Transformer-based prediction module, the Seasonal-Trend Decomposition module and the Decomposition Adaptation module. The core idea of SeDAN is that in the knowledge transfer process, we do not directly align the features obtained from the encoder, but align the feature components decomposed from the original encoder features. Due to the low coupling between the feature components, we can apply corresponding adaptation methods to the components respectively. Specifically, for the source domain data $X^s$ and the target domain data $X^t$, the basic prediction module performs feature encoding to obtain original feature sequences $H^s$ and $H^t$. Then we design ICD to conduct the Season-Trend Decomposition for original feature sequences. For $H^s$, we decompose it into the seasonal feature component $H_{sea}^ s$ and trend feature component $H_{tre}^s$. Similarly, for $H^t$, we also decompose it into $H_{sea}^t$ and $H_{tre}^t$. For the source and target domains, we perform adaptation of the two types of features respectively. We use joint probability adaptation to define the seasonal feature metric $\mathscr{D}^s$, by minimizing the distance $<\phi ^u(H_{sea}^s) , \phi ^u(H_{sea }^t)>^{\mathscr{D}^s}$ in the general feature space $\phi ^u$. And we propose an Optimal Local Adaptation as the trend feature metric $\mathscr{D}^t$ , to minimize the distance $<\phi ^u(H_{tre}^s) , \phi ^u(H_{tre}^t)>^{\mathscr{D}^t}$. The knowledge transfer process of SeDAN is mainly reflected in the Seasonal-Trend Decomposition module and the Decomposition Adaptation module. After the knowledge transfer process, we reconstruct the original encoder features using the seasonal and trend features, and use the reconstructed $ \tilde{H}^s$ and $ \tilde{H}^t$ for prediction.

The basic prediction model of SeDAN is built on the encoder-decoder framework. The encoder generates the corresponding feature sequence $H = ( h_1, h_2, \cdots, h_t )$ for the input time series $X$ , while the reconstructed $H$ is the input of the decoder to generate prediction results $\hat{Y}$. In this paper, we use vanilla Transformer to construct our model. We utilize stacked self-attention structure [14] and position-wise feed forward network as encoder, and utilize masked self-attention as decoder. We use Local Time Stamp and Global Time Stamp \citep{15-informer-DBLP:conf/aaai/ZhouZPZLXZ21} as Positional Encoding.

In the learning process of SeDAN, the objective function consists of three parts. We define MSE as the objective function of basic prediction learning, define infoNCE Loss for decomposition learning, and define $\mathscr{D}^s$ and $\mathscr{D }^t$ for feature adaptation. Based on the above items, the final objective function of SeDAN can be formalized as follows:

\begin{equation}
\mathcal{L}=\underbrace{\mathcal{L}_{mse}\left(\hat{\mathbf{Y}}, \mathbf{Y}\right)}_{\text {forecasting }} \\
+\lambda\underbrace{\mathcal{L}_{dec}\left(\mathbf{H}^{s}, \mathbf{H}^{t}\right)}_{\text {decomposition }} \\
+\gamma\underbrace{\mathcal{L}_{\mathscr{D}^s,\mathscr{D}^t}\left(\mathbf{H}_{sea}, \mathbf{H}_{tre}\right)}_{\text {adaptation }},
\end{equation}

\noindent where $\lambda$ represents the decomposition ratio and $\gamma$ represents the adaption ratio, both of which are hyper parameters.

During inference, we only keep the neural networks for the target domain data.

\subsection{Implicit Contrastive Decomposition}
\label{ssec:4-2}

In this subsection, we introduce the proposed ICD for the Seasonal-Trend Decomposition. In this module, the feature sequence extracted by encoder is decomposed into low-coupling seasonal and trend features for follow-up decomposition adaptation. The original feature sequence $H$ contains the entangled seasonal feature $H_{sea}$ and the trend feature $H_{tre}$. In order to obtain $H_{sea}$ and $H_{tre}$, we design Seasonal Decomposition Generator (SDG) and Trend Decomposition Generator (TDG). 
It is necessary to find an efficient way to make the generated features have periodic properties and trend properties respectively. Besides, the model should be able to reconstruct the original feature sequence with the two generated features in order to prevent information loss.

Since the seasonal and trend features are difficult to define explicitly, inspired by contrastive learning \citep{51-CL-survey-DBLP:journals/corr/abs-2006-08218,52-CL-DBLP:conf/cvpr/He0WXG20,53-CL-DBLP:conf/icml/ChenK0H20}, we design an Instance-Instance Contrast method called ICD. The method captures the periodic property invariance and trend property invariance between instances to construct the self-supervised objective functions for SDG and TDG. Specifically, we construct a dictionary look-up task. For the $H$ as input, the positive samples $K_i^{+}$ are generated by data augmentation that maintains the periodicity and trend of the input sequence. And the negative samples $K_i$ are sampled from other sequences in the mini-batch. We put $H$ pass through the data augmentation and the generators to construct the query of the look-up task, and regard the $K_i^{+}$ and $K_i$ as keys, so as to construct a contrastive learning task for SDG and TDG. In order to improve the effectiveness of contrastive learning, we construct a memory bank with negative samples by a queue. In order to ensure the consistency of negative samples in the memory bank, we refer to MoCo [52] and use a momentum encoder to generate keys. Different from MoCo, when a new batch of data needs to enter the queue, we no longer use a simple first-in-first-out method, but use an Online Prototype Update (OPU) mechanism. The time series dataset is relatively small, and directly discarding the first-in samples leads to a decrease in the richness of negative samples. Therefore, we aim to use the samples effectively through OPU. For $K_i$ which will be discarded from the queue, OPU selects the most similar sample $K_j$ from the existing keys of memory bank as the prototype, and then fuse $K_i$ and $K_j$ to update the prototype, leading to a more representative prototype. The calculation of OPU can be formulized as follows:

\begin{equation}
\begin{aligned}
\mathop{\arg\min}_{K_j} &= \frac{<K_i , K_j>^{\mathscr{D}^{opu}}}{\sum_{j=1}^{N_{b}} <K_i , K_j>^{\mathscr{D}^{opu}}}, \\
K_j &= (1- \beta)K_j + \beta K_i,
\end{aligned}
\end{equation}

\noindent where $N_b$ is the capacity of the memory bank and $\beta$ is the update ratio. We use the dot product as $\mathscr{D}^{opu}$. 

Then we introduce the decomposition generators of seasonal and trend features in detail.

\paragraph{Seasonal feature decomposition}
We define MLP as the SDG to generate the seasonal feature $H_{sea}$:

\begin{equation}
H_{sea} = MLP^{SDG}(H).
\end{equation}

We leverage the above method to provide supervision for SDG. In the process of constructing positive samples, we choose data augmentation methods that can maintain the periodicity of input sequence, including rolling, flipping, and scaling \citep{59-TS-DA-survey-DBLP:conf/ijcai/Wen0YSGWX21}. For the input feature sequence $H$, the relevant methods are specifically defined as follows:

\begin{itemize}
    \item Rolling: scroll the sequence on the time axis. If we move the sequence $h$ for $r$ time steps, the items in new sequence are $h_i = h_{i - Sgn(i-r) \cdot r}$.
    
    \item Flipping: flip the sequence horizontally, and the items in new sequence are $h_i = h_{m-i-1} $.
    
    \item Scaling: scale the sequence elements, and the items in new sequence are $h_i = \epsilon_s \cdot h_i$.
\end{itemize}

The above data augmentation methods do not alter the inherent periodicity of the sequence. We use a combination of these methods to generate the query and positive samples of seasonal feature:

\begin{equation}
\begin{aligned}
Q^{sea} &= MLP^{SDG}(f^{sea}_{da}(H)), \\
K^{sea+} &= MLP^{Moco}(f^{sea}_{da}(H_i)),
\end{aligned}
\end{equation}

\noindent where $f^{sea}_{da}$ represents the combined data augmentation process for seasonal decomposition. $MLP^{Moco}$ and $MLP^{SDG}$  have the same structure but momentum update is used for $MLP^{Moco}$ . At the same time, we sample $k$ negative samples from the memory bank. For positive samples $K^{sea+}$ and the negative samples $\{ K_1^{sea}, K_2^{sea}, \cdots, K_k^{sea}\}$, we use infoNCE Loss as the contrastive loss function:

\begin{equation}
\mathcal{L}_{dec}^{sea}=-\log \frac{\exp \left(Q^{sea} \cdot K^{sea+} / \tau\right)}{\exp \left(Q^{sea} \cdot K^{sea+} / \tau\right)+\sum_{j=1}^{k} \exp \left(Q^{sea} \cdot K_{j}^{sea} / \tau\right)},
\end{equation}

\noindent where $\tau$ is the temperature hyper parameter. To enrich the diversity of keys, we randomly combine the above three data augmentation methods to generate positive samples.

\paragraph{Trend feature decomposition}
For the TDG, inspired by the STL \citep{47-decomposition-STL-cleveland1990stl}, we use the causal convolution \citep{13-TCN-DBLP:journals/corr/abs-1803-01271} and average pooling to simulate the trend smoothing process. The generation process of the trend feature $H_{tre}$ is formulized as:

\begin{equation}
H_{tar} = AvgPool(Conv_{causal}^{TDG}(H)).
\end{equation}

Similar to SDG, we also construct a contrastive method to provide supervision for TDG. We choose the data augmentation methods that maintain the trend of the sample sequence to construct positive samples, including jittering, window cropping, and window warping \citep{59-TS-DA-survey-DBLP:conf/ijcai/Wen0YSGWX21}. The specific definitions are as follows:

\begin{itemize}
    \item Jittering: add random noise to the sequence, and the the items in new sequence are $h_i = h_i + \epsilon_j^i $, where $\epsilon_j^i \sim \mathcal{N}(0, \sigma^2_j)$.
    
    \item Window cropping: randomly extract consecutive slices from the original sequence as the augmented sequence.
    
    \item Window warping: choose a random time range of the original sequence and then compress or expand while keeping the other time ranges unchanged.
\end{itemize}

For positive samples $K^{tre+}$ and negative samples $\{ K_1^{tre}, K_2^{tre}, \cdots, K_k^{tre}\}$, we also use infoNCE as a contrastive loss function $\mathcal {L}^{tre}_{dec}$, which formula has a similar form to $\mathcal{L}_{dec}^{sea}$ .

After generating the decomposed components $H_{sea}$ and $H_{tre}$, we reconstruct the components as $\tilde{H}$ using the reconstructor and KL divergence. We hope $\tilde{H}$ to keep the information contained in the original feature sequence $H$ as much as possible. Therefore, we can obtain the complete objective function of the ICD as:

\begin{equation}
\mathcal{L}_{dec} = \mathcal{L}_{dec}^{sea} + \mathcal{L}_{dec}^{tre} + \mathcal{L}_{KL}( H || \tilde{H}).
\end{equation}

It should be noted that since we use implicit decomposition generation instead of explicit decomposition \citep{47-decomposition-STL-cleveland1990stl,48-decomposition-EMD-huang1998empirical,49-decomposition-X11-shiskin1967x}, the decomposed seasonal and trend features no longer satisfy the additive or multiplicative model. The reconstruction helps SeDAN to reduce information loss during decomposition.

\subsection{Decomposition Adaptation}
\label{ssec:4-3}

The Decomposition Adaptation is carried out under the assumption that the marginal distributions of different time series are transferable and the conditional distributions are partly transferable. Because the seasonal features reflect the periodicity of the sequence, the stochastic transition dynamics is relatively fixed, which can be transferred. In contrast, the transition dynamics in the trend features is more related to the domain characteristic, and directly transferring the trend features may lead to negative transfer. Based on the above reasons, we argue that the seasonal features include the marginal distribution and the transferable conditional distribution of the sequence, while the trend features include the marginal distribution and the non-transferable conditional distribution. Under this assumption, we design different adaptation methods for seasonal and trend features separately, which aim to only adapt the transferable part in different domains. We introduce the adaptation methods for the two types of features respectively in subsection~\ref{sssec:4-3-1} and \ref{sssec:4-3-2}.

\subsubsection{Seasonal Feature Adaptation}
\label{sssec:4-3-1}

The Seasonal Feature Adaptation is to align the seasonal features $H_{sea}^s$ in source domain and $H_{sea}^t$ in target domain. We argue that subsequences can better reflect the characteristic patterns in sequence data than a single time step. Through observation, it can be found that subsequences in the same or different time series often have similar morphology in the time domain. The powerful performance of the forecasting algorithms based on the sliding window also shows the rationality of this assumption. More recently, \citeauthor{61-AdaRNN-DBLP:conf/cikm/Du0FPQXW21} formalize this distribution assumption, defining it as the Temporal Covariate Shift. In our work, we use the subsequence-based distribution assumption.

For the seasonal feature sequence, such as $H_{sea}^t$ in the target domain, we split it into $n_t$ subsequences: $H_{sea}^t = \{Se^t_1, Se^t_2 , \cdots, Se^t_{n_t} \}$. We consider that each subsequence $Se_i^t$ reflects the marginal distribution in the sequence, and the conversion method between consecutive subsequences reflects the conditional distribution. Similarly for the source domain, we also have $H_{sea}^s = \{Se^s_1, Se^s_2, \cdots, Se^s_{n_s} \}$ with $n_s$ subsequences. 
Since the distribution contained in seasonal features is fully transferable, we can adapt the two distributions at the same time. Therefore, we define the joint distribution adaptation as the adaptation method of $H_{sea}^s$ and $H_{sea}^t$. Considering the properties of time series data, we use the first-order markov property to simplify the Joint Maximum Mean Discrepancy (JMMD) [62] as the seasonal metric $\mathscr{D}^s$. The adaptation metric of $H_{sea}^s$ and $H_{sea}^t$ in the feature space can be calculated as follows:

\begin{equation}
\begin{aligned}
\mathscr{D}^s (H_{sea}^s, H_{sea}^t ) &= \frac{1}{n_{s}} \sum_{i=1}^{n_{s}} k_e \left(Se_{i}^{s}, Se_{j}^{s}\right) \\
&+ \frac{1}{n_{t}} \sum_{i=1}^{n_{t}} k_e \left(Se_{i}^{t}, Se_{j}^{t}\right) \\
&+\frac{2}{n_{s}n_{t}} \sum_{i=1}^{n_{s}} \sum_{j=1}^{n_{t}} k_e \left(Se_{i}^{s}, Se_{j}^{t}\right),
\end{aligned}
\end{equation}

\noindent where $k_e$ is the kernel function used to regenerate the inner product transformation in the kernel hilbert space, and we use gaussian kernel in this paper. The first-order markov property effectively reduces the computational complexity of JMMD.


\begin{figure*}[htp]
\centering
\includegraphics[width=0.75\textwidth]{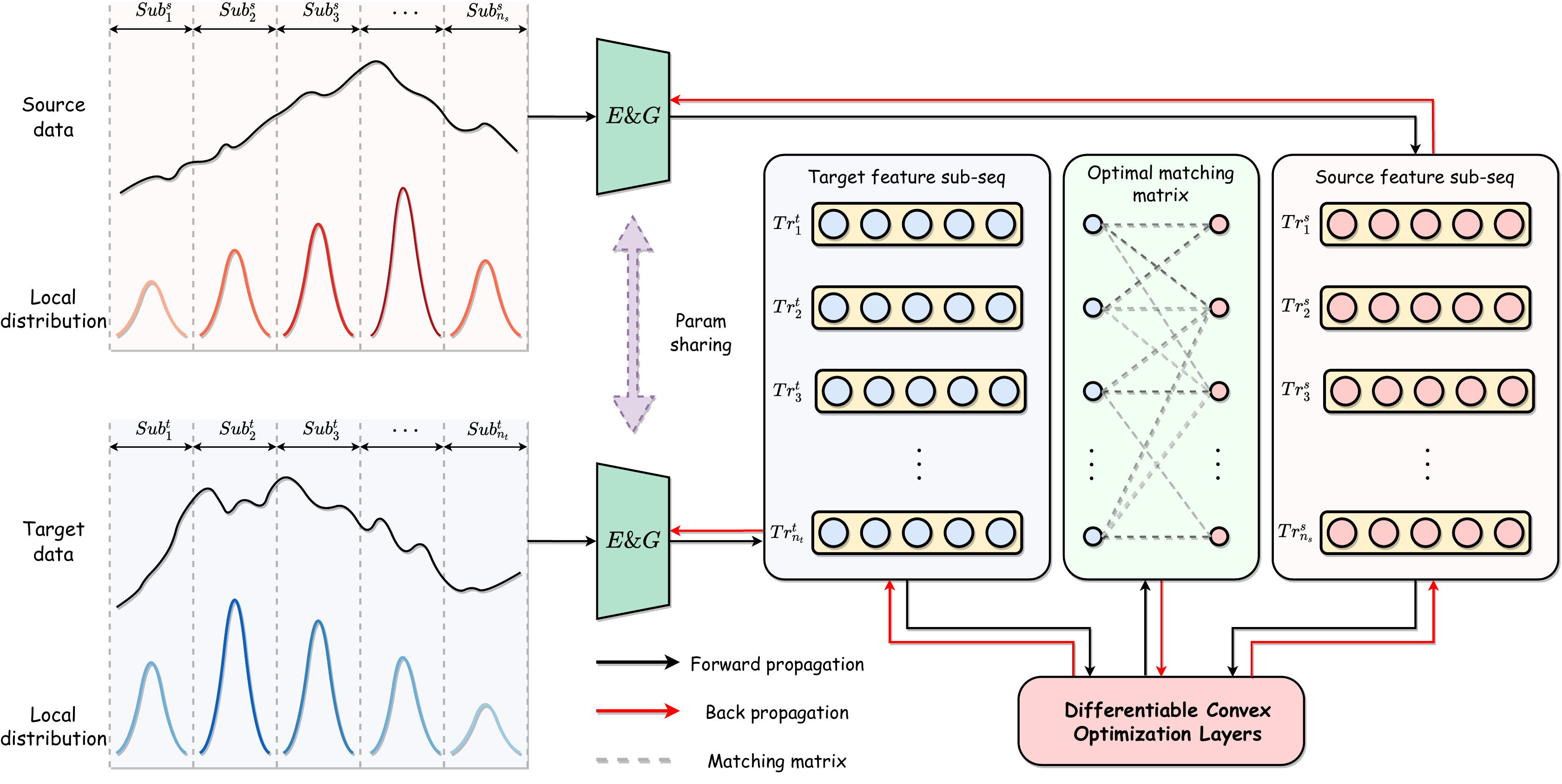}\par
\caption{The proposed OLA adaptation module. Following the subsequence-based distribution assumption, the trend feature adaptation is achieved by matching subsequences with similar distributions in the source and target domains.}
\label{fig:fig-4}
\end{figure*}

\subsubsection{Trend Feature Adaptation}
\label{sssec:4-3-2}

The Trend Feature Adaptation is to align $H_{tre}^s$ and $H_{tre}^t$. Different from the Seasonal Feature Adaption, the conditional distributions contained in $H_{tre}^s$ and $H_{tre}^t$ are more difficult to transfer. Only the marginal distributions can be transferred to help the model learn domain-independent knowledge. Because the subsequences with similar features may have different indices in the two sequences, it is unreasonable to adapt the marginal distribution directly to the sequential $H_{tre}^s$ and $H_{tre}^t$, which can be formulized as $\sum_{i=1}^{t}\mathscr{D}^ t(h_i^s, h_i^t)$. If the order measurement based on the global representation is directly performed, the model will discard the local features in the sequence when learning domain-independent feature representations, which are critical for time series forecasting. Therefore, we propose OLA to deal with the problem that similar subsequences have different indices so as to retain meaningful local characteristics during the transfer process.

The OLA also follows the subsequence-based distribution assumption, and each subsequence is regarded as a local characteristic in the complete sequence. OLA formulates the adaptation problem as a global optimal matching problem and solves it using a linear programming method. In the global scope, similar matching subsequences are found for two sets of feature subsequences in the source domain and the target domain to achieve trend feature adaptation. Specifically, for $H_{tre}^s = \{Tr^s_1, Tr^s_2, \cdots, Tr^s_{n_s} \}$ and $H_{tre}^t= \{Tr^ t_1, Tr^t_2, \cdots, Tr^t_{n_t} \}$, we define the cost matrix $ C \in \mathbb{R}^{n_s \times n_t }$, where $c_i^ j$ represents the matching cost between subsequences $Tr^s_i$ and $Tr^t_i$. We hope to find a matching matrix $M \in \mathbb{R}^{n_s \times n_t }$, so that the matched two sets of subsequences have the minimum matching cost under constraints. Because we aim to find the similar match between two sets of subsequences, we can define the cost matrix $C = A  - S$, where $A= \{ a_{ij} = 1 | i = 1,2 , \cdots n_s , j = 1,2, \cdots n_t \} $, and $ S \in \mathbb{R}^{n_s \times n_t }$ is the similarity matrix between the two sets of subsequences. We formalize the optimal matching problem and the constraints as follows:

\begin{equation}
\begin{aligned}
\underset{m_{i j}}{\operatorname{minimize}} & \sum_{i=1}^{n_s} \sum_{j=1}^{n_t} (1 - s_{i j}) m_{i j},  \\
\text { s.t. } & m_{i j} \geq 0, \quad i=1, \cdots, n_s, j=1, \cdots, n_t, \\
& \sum_{i=1}^{n_s} m_{i j}=1, \quad j=1, \ldots n_t, \\
& \sum_{j=1}^{n_t} m_{i j}=1, \quad i=1, \ldots n_s.
\end{aligned}
\end{equation}

Cosine similarity is used to define the similarity matrix $S$:

\begin{equation}
s_{ij} = \frac{Tr^s_i \odot Tr^t_j} {\|Tr^s_i\| \times\left\|Tr^t_i\right\|},
\end{equation}

\noindent where $\odot$ represents element-wise multiplicatio between matrices. The optimal matching problem is a standard Linear Programming problem whose Lagrangian can be expressed as:

\begin{equation}
L(M, \nu, \lambda)=(A-S)^{T} M+ \nu^{T}(A M-b) +\eta^{T}(G M-h),
\end{equation}

\noindent where $A$ and $b$ represent the equality constraints in optimization problems. $G$ and $h$ represent the inequality constraints. $\nu$ and $\eta$ are dual variables for equality and inequality constraints. We choose to use the primal-dual interior point method to solve this convex optimization problem. The OLA is used as the trend metric $\mathscr{D}^t$. Having the optimal matching matrix $\tilde{M}$, the adaption for $H_{tre}^s$ and $H_{tre}^t$ is calculated as follows:

\begin{equation}
\mathscr{D}^t(H_{tre}^s, H_{tre}^t) = (A-S)\odot \tilde{M}.
\end{equation}

The optimization problem should be embed into the network to optimize the parameters $\theta$ for learning. Therefore, the Differentiable Convex Optimization Layers is introduced to perform differentiable processing on this optimization problem, so that the gradient can be back-propagated on the optimization layer. The overall OLA adaptation module is shown in Figure~\ref{fig:fig-4}. We combine $\mathscr{D}^s$ and $\mathscr{D}^t$ to construct the complete objective function of the adaptation:

\begin{equation}
\mathcal{L}_{\mathscr{D}^s,\mathscr{D}^t}\left(\mathbf{H}_{sea}, \mathbf{H}_{tre}\right) = \mathscr{D}^s(H_{sea}^s, H_{sea}^t) + \mathscr{D}^t(H_{tre}^s, H_{tre}^t).
\end{equation}

Through decomposition and adaptation, we complete the alignment of source and target domain data in the feature space.

\section{Experiments}
\label{sec:experiments}

We validate the performance of our SeDAN on multiple multivariate time series forecasting datasets. In subsection~\ref{ssec:5-1}, we introduce the datasets and the implementation details of the experiments. In subsection~\ref{ssec:5-2}, we introduce the comparison methods including the single-domain baselines and the cross-domain baselines. In subsection~\ref{ssec:5-3}, we report the results and provide detailed analysis of the experiments. In subsection~\ref{ssec:5-4} and \ref{ssec:5-5}, we conduct ablation experiments and further analyze the properties of our proposed model.

\begin{table*}[htbp]
\centering
\caption{Performance comparison with single-domain methods on five target datasets.}
\label{table:table-1}
\scalebox{1}{
\begin{tabular}{cccccccccccccc}
\toprule
\multicolumn{2}{c}{Models}                                             & \multicolumn{2}{c}{SeDAN(Best)} & \multicolumn{2}{c}{SeDAN(Avg)} & \multicolumn{2}{c}{Informer} & \multicolumn{2}{c}{TCN} & \multicolumn{2}{c}{LSTNet} & \multicolumn{2}{c}{LSTMa} \\ 
\cmidrule(lr){3-4}\cmidrule(lr){5-6}\cmidrule(lr){7-8}\cmidrule(lr){9-10}\cmidrule(lr){11-12}\cmidrule(lr){13-14}
\multicolumn{2}{c}{Metric}                                             & MSE            & MAE            & MSE            & MAE           & MSE           & MAE          & MSE        & MAE        & MSE          & MAE         & MSE         & MAE         \\ \midrule
\multicolumn{1}{c|}{\multirow{3}{*}{ETTh1}} & \multicolumn{1}{c|}{24}  & \textbf{0.491} & \textbf{0.499} & 0.509          & 0.509         & 0.577         & 0.549        & 0.583      & 0.547      & 1.293        & 0.901       & 0.650        & 0.624       \\
\multicolumn{1}{c|}{}                       & \multicolumn{1}{c|}{48}  & \textbf{0.592} & \textbf{0.555} & 0.611          & 0.567         & 0.685         & 0.625        & 0.670       & 0.606      & 1.456        & 0.960        & 0.702       & 0.675       \\
\multicolumn{1}{c|}{}                       & \multicolumn{1}{c|}{168} & {0.827} & {0.688} & 0.834          & 0.693         & 0.931         & 0.752        & \textbf{0.811}      & \textbf{0.680}       & 1.997        & 1.214       & 1.212       & 0.867       \\ \midrule
\multicolumn{1}{c|}{\multirow{3}{*}{ETTh2}} & \multicolumn{1}{c|}{24}  & \textbf{0.389} & \textbf{0.480}  & 0.425          & 0.502         & 0.720          & 0.665        & 0.935      & 0.754      & 2.742        & 1.457       & 1.143       & 0.813       \\
\multicolumn{1}{c|}{}                       & \multicolumn{1}{c|}{48}  & \textbf{0.746} & \textbf{0.700} & 0.788          & 0.715         & 1.457         & 1.001        & 1.300      & 0.911      & 3.567        & 1.687       & 1.671       & 1.221       \\
\multicolumn{1}{c|}{}                       & \multicolumn{1}{c|}{168} & \textbf{1.691} & \textbf{1.013} & 1.737          & 1.037         & 3.489         & 1.515        & 4.017      & 1.579      & 3.242        & 2.513       & 4.117       & 1.674       \\ \midrule
\multicolumn{1}{c|}{\multirow{3}{*}{WTH}}   & \multicolumn{1}{c|}{24}  & \textbf{0.316} & \textbf{0.340} & 0.312          & 0.354         & 0.335         & 0.381        & 0.321      & 0.367      & 0.615        & 0.545       & 0.546       & 0.570       \\
\multicolumn{1}{c|}{}                       & \multicolumn{1}{c|}{48}  & \textbf{0.369} & \textbf{0.409} & 0.373          & 0.414         & 0.395         & 0.459        & 0.386      & 0.423      & 0.660        & 0.589       & 0.829       & 0.677       \\
\multicolumn{1}{c|}{}                       & \multicolumn{1}{c|}{168} & {0.503} & {0.502} & 0.503          & 0.504         & 0.608         & 0.567        & \textbf{0.491}     & \textbf{0.501}      & 0.748        & 0.647       & 1.038       & 0.835       \\ \midrule
\multicolumn{1}{c|}{\multirow{3}{*}{EXC}}   & \multicolumn{1}{c|}{24}  & \textbf{0.166} & \textbf{0.315} & 0.187          & 0.343         & 0.338         & 0.460        & 0.499      & 0.452      & 0.576        & 0.581       & 0.504       & 0.440       \\
\multicolumn{1}{c|}{}                       & \multicolumn{1}{c|}{48}  & \textbf{0.455} & \textbf{0.541} & 0.541          & 0.600         & 0.847         & 0.752        & 1.759      & 1.130      & 1.551        & 1.058       & 1.473       & 0.946       \\
\multicolumn{1}{c|}{}                       & \multicolumn{1}{c|}{168} & \textbf{0.726} & \textbf{0.688} & 0.735          & 0.695         & 0.972         & 0.766        & 2.462      & 1.659      & 1.822        & 1.212       & 1.776       & 1.181       \\ \midrule
\multicolumn{1}{c|}{\multirow{3}{*}{ILI}}   & \multicolumn{1}{c|}{24}  & \textbf{3.196} & \textbf{1.230} & 3.335          & 1.268         & 5.764         & 1.677        & 6.624      & 1.830      & 6.026        & 1.770       & 5.924       & 1.710       \\
\multicolumn{1}{c|}{}                       & \multicolumn{1}{c|}{36}  & \textbf{3.069} & \textbf{1.186} & 3.284          & 1.223         & 4.755         & 1.467        & 6.858      & 1.879      & 5.340         & 1.668       & 6.416       & 1.794       \\
\multicolumn{1}{c|}{}                       & \multicolumn{1}{c|}{60}  & \textbf{3.710}  & \textbf{1.298} & 3.807          & 1.333         & 5.264         & 1.564        & 7.127      & 1.918      & 5.548        & 1.720        & 6.736       & 1.833       \\ \bottomrule
\end{tabular}
}
\end{table*}

\subsection{Datasets and implementations}
\label{ssec:5-1}

\subsubsection{Datasets}
\label{sssec:5-1-1}

We extensively evaluate the proposed SeDAN on 5 real-world benchmark datasets.

\begin{itemize}
    \item ETT\footnote{https://github.com/zhouhaoyi/ETDataset} (Electricity Transformer Temperature) \citep{15-informer-DBLP:conf/aaai/ZhouZPZLXZ21}: The ETT dataset collects electricity transformer data from two different counties in China between July 2016 and July 2018, including the oil temperature and six other electricity load features. ETT consists of two 1-hour-level datasets \{ETTh1, ETTh2\} and two 15-minute-level datasets \{ETTm1, ETTm2\}. Only ETTh1 and ETTh2 are used in our experiments.
    
    \item Weather\footnote{https://www.ncei.noaa.gov/data/local-climatological-data/}: The Weather dataset collects local climate data from nearly 1,600 U.S. locations between 2010 and 2013, including the wet bulb and the remaining 11 climate features. The data points are collected hourly.
    
    \item Exchange-Rate\citep{32-RNN-LSTNet-DBLP:conf/sigir/LaiCYL18}: The Exchange-Rate dataset collects daily exchange rates for eight different countries from 1990 to 2016.
    
    \item ILI\footnote{https://gis.cdc.gov/grasp/fluview/fluportaldashboard.html} (Inﬂuenza-like Illness)\citep{67-Autoformer-DBLP:conf/nips/WuXWL21}: The ILI dataset collects weekly data on inﬂuenza-like illness (ILI) patients in the United States from 2002 to 2021, including seven features such as the proportion and total number of ILI patients.
\end{itemize}

To verify the effectiveness of the proposed model, we set up single-domain and cross-domain comparative experiments respectively. In the single-domain experiment, the compared models only use the target domain dataset for training and inference. In the cross-domain experiment, we choose the above datasets as the target datasets and set at least three source datasets for each target dataset separately, which follows the rule that the size of the source dataset is roughly equal or larger than the size of the target dataset. Due to the large scale of data in the WTH dataset, we choose a larger ECL dataset \citep{15-informer-DBLP:conf/aaai/ZhouZPZLXZ21} as one of its source datasets, which collects the hourly electricity consumption of 321 customers from 2012 to 2014.

\begin{table*}[htb]
\centering
\caption{Performance comparison with cross-domain methods on five target datasets, each with three source datasets.}
\label{table:table-2}
\scalebox{1}{
\begin{tabular}{ccccccccccccc}
\toprule
\multicolumn{3}{c}{Model}                                                                        & \multicolumn{2}{c}{SeDAN}         & \multicolumn{2}{c}{DAN} & \multicolumn{2}{c}{JAN} & \multicolumn{2}{c}{Fine-tuning} & \multicolumn{2}{c}{vanilla Trx}                                                                           \\ 
\cmidrule(lr){1-3}\cmidrule(lr){4-5}\cmidrule(lr){6-7}\cmidrule(lr){8-9}\cmidrule(lr){10-11}\cmidrule(lr){12-13} Target                                      & Length                                    & Source & MSE             & MAE             & MSE    & MAE            & MSE        & MAE        & MSE            & MAE            & MSE                                                 & MAE                                                 \\ 
\midrule \multicolumn{1}{c|}{\multirow{6}{*}{ETTh1}} & \multicolumn{1}{c|}{\multirow{3}{*}{24}}  & ETTh2  & \textbf{0.512}  & \textbf{0.510}  & 0.568  & 0.521          & 0.572      & 0.522       & 0.550          & 0.525          & \multirow{3}{*}{0.583}                                   & \multirow{3}{*}{0.556}                             \\
\multicolumn{1}{c|}{}                       & \multicolumn{1}{c|}{}                     & WTH    & \textbf{0.491}  & \textbf{0.499}  & 0.559  & 0.522          & 0.531      & 0.501      & 0.501         & \textbf{0.499}                                                    \\ 
\multicolumn{1}{c|}{}                       & \multicolumn{1}{c|}{}                     & ECL    & \textbf{0.522}  & \textbf{0.516}  & 0.555  & 0.535          & 0.576      & 0.523      & 0.564          & 0.556                                                    \\ 
\cmidrule{2-13}  \multicolumn{1}{c|}{}                       & \multicolumn{1}{c|}{\multirow{3}{*}{168}} & ETTh2  & \textbf{0.827} & \textbf{0.688}  & 0.909  & 0.711          & 0.894      & 0.696      & 0.923          & 0.729          & \multirow{3}{*}{0.949}                              & \multirow{3}{*}{0.762}                              \\ 
\multicolumn{1}{c|}{}                       & \multicolumn{1}{c|}{}                     & WTH    & \textbf{0.843}  & \textbf{0.698}  & 0.882  & 0.703          & 0.901      & 0.707      & 0.952          & 0.750                                                    \\
\multicolumn{1}{c|}{}                       & \multicolumn{1}{c|}{}                     & ECL    & \textbf{0.830}  & \textbf{0.690}  & 0.969   & 0.753          & 0.875      & 0.699      & 0.929          & 0.736                                                    \\ 
\midrule \multicolumn{1}{c|}{\multirow{6}{*}{ETTh2}} & \multicolumn{1}{c|}{\multirow{3}{*}{24}}  & ETTh1  & \textbf{0.389}  & \textbf{0.480}   & 0.517  & 0.571          & 0.538      & 0.568      & 0.496         & 0.549         & \multirow{3}{*}{0.747}                              & \multirow{3}{*}{0.694}                              \\
\multicolumn{1}{c|}{}                       & \multicolumn{1}{c|}{}                     & WTH    & \textbf{0.461}  & \textbf{0.529}  & 0.467  & 0.548          & 0.510      & 0.550      & 0.555          & 0.577                                                 \\
\multicolumn{1}{c|}{}                       & \multicolumn{1}{c|}{}                     & ECL    & \textbf{0.426}  & \textbf{0.497} & 0.611  & 0.640           & 0.588      & 0.597      & 0.614          & 0.606                                                     \\ 
\cmidrule{2-13}  \multicolumn{1}{c|}{}                       & \multicolumn{1}{c|}{\multirow{3}{*}{168}} & ETTh1  & \textbf{1.772}  & \textbf{1.092}  & 2.853  & 1.413          & 2.089      & 1.168      & 3.794          & 1.641          & \multicolumn{1}{l}{\multirow{3}{*}{{3.569}}} & \multicolumn{1}{l}{\multirow{3}{*}{{1.516}}} \\
\multicolumn{1}{c|}{}                       & \multicolumn{1}{c|}{}                     & WTH    & \textbf{1.748}  & \textbf{1.007}  & 2.275  & 1.233          & 1.834      & 1.042      & 2.678          & 1.287          & \multicolumn{1}{l}{}                                & \multicolumn{1}{l}{}                                \\
\multicolumn{1}{c|}{}                       & \multicolumn{1}{c|}{}                     & ECL    & \textbf{1.691}  & \textbf{1.013}  & 3.129  & 1.485          & 2.176      & 1.167      & 3.396          & 1.528          & \multicolumn{1}{l}{}                                & \multicolumn{1}{l}{}                                \\ 
\midrule \multicolumn{1}{c|}{\multirow{6}{*}{WTH}}   & \multicolumn{1}{c|}{\multirow{3}{*}{24}}  & ETTh1  & \textbf{0.316}  & \textbf{0.340}  & 0.320  & 0.370          & 0.334      & 0.360      & 0.340          & 0.397          & \multirow{3}{*}{0.348}                              & \multirow{3}{*}{0.398}                              \\
\multicolumn{1}{c|}{}                       & \multicolumn{1}{c|}{}                     & ETTh2  & \textbf{0.312}  & \textbf{0.364}  & 0.338  & 0.387          & 0.340      & 0.369      & 0.332          & 0.383                                                  \\
\multicolumn{1}{c|}{}                       & \multicolumn{1}{c|}{}                     & ECL    & \textbf{0.308}  & \textbf{0.359}  & 0.322  & 0.373          & 0.346      & 0.367      & 0.330          & 0.383                                                     \\ 
\cmidrule{2-13}  \multicolumn{1}{c|}{}                       & \multicolumn{1}{c|}{\multirow{3}{*}{168}} & ETTh1  & \textbf{0.506}  & \textbf{0.504}  & 0.543  & 0.541          & 0.555      & 0.546      & 0.559          & 0.562          & \multirow{3}{*}{0.617}                              & \multirow{3}{*}{0.572}                              \\
\multicolumn{1}{c|}{}                       & \multicolumn{1}{c|}{}                     & ETTh2  & \textbf{0.503}  & \textbf{0.502}  & 0.539  & 0.539          & 0.557      & 0.549      & 0.532          & 0.530                                                     \\
\multicolumn{1}{c|}{}                       & \multicolumn{1}{c|}{}                     & ECL    & \textbf{0.499}  & \textbf{0.507}  & 0.580  & 0.577          & 0.561      & 0.556      & 0.549          & 0.549                                                     \\ 
\midrule \multicolumn{1}{c|}{\multirow{6}{*}{EXC}}   & \multicolumn{1}{c|}{\multirow{3}{*}{24}}  & ETTh1  & \textbf{0.166}  & \textbf{0.315}  & 0.355  & 0.454          & 0.248      & 0.408      & 0.211          & 0.369          & \multirow{3}{*}{0.341}                              & \multirow{3}{*}{0.464}                              \\
\multicolumn{1}{c|}{}                       & \multicolumn{1}{c|}{}                     & ETTh2  & \textbf{0.207}  & \textbf{0.365}  & 0.322 & 0.426         & 0.283     & 0.431     & 0.261          & 0.397                                                    \\
\multicolumn{1}{c|}{}                       & \multicolumn{1}{c|}{}                     & WTH    & \textbf{0.188}  & \textbf{0.350}   & 0.265  & 0.399          & 0.213      & 0.374      & 0.211          & 0.361                                                             \\ 
\cmidrule{2-13}  \multicolumn{1}{c|}{}                       & \multicolumn{1}{c|}{\multirow{3}{*}{168}} & ETTh1  & \textbf{0.728}  & \textbf{0.694}  & 0.996  & 0.738          & 0.775      & 0.707      & 0.757          & 0.716          & \multirow{3}{*}{0.981}                              & \multirow{3}{*}{0.781}                              \\
\multicolumn{1}{c|}{}                       & \multicolumn{1}{c|}{}                     & ETTh2  & \textbf{0.726}  & \textbf{0.688}  & 0.806  & 0.703          & 0.770      & 0.710      & 0.971          & 0.742                                                    \\
\multicolumn{1}{c|}{}                       & \multicolumn{1}{c|}{}                     & WTH    & \textbf{0.750}   & \textbf{0.702}  & 0.888  & 0.711          & 0.887      & 0.745      & 1.100          & 0.770                                                     \\ 
\midrule \multicolumn{1}{c|}{\multirow{6}{*}{ILI}}   & \multicolumn{1}{c|}{\multirow{3}{*}{24}}  & ETTh1  & \textbf{3.196}  & 1.230            & 4.649  & 1.435          & 3.597      & \textbf{1.214}      & 5.351          & 1.542          & \multirow{3}{*}{5.821}                              & \multirow{3}{*}{1.687}                              \\
\multicolumn{1}{c|}{}                       & \multicolumn{1}{c|}{}                     & WTH    & \textbf{3.480}   & 1.287           & 3.621  & \textbf{1.277} & 4.198      & 1.427      & 5.743          & 1.630                                                            \\
\multicolumn{1}{c|}{}                       & \multicolumn{1}{c|}{}                     & ECL    & \textbf{3.330}  & 1.286  & 4.172  & 1.399          & 3.675      & \textbf{1.245}      & 5.377          & 1.580                                                   \\ 
\cmidrule{2-13}  \multicolumn{1}{c|}{}                       & \multicolumn{1}{c|}{\multirow{3}{*}{60}}  & ETTh1  & \textbf{3.710}  & \textbf{1.298}  & 4.193  & 1.413          & 4.727      & 1.433      & 4.825          & 1.415          & \multirow{3}{*}{5.311}                              & \multirow{3}{*}{1.581}                              \\
\multicolumn{1}{c|}{}                       & \multicolumn{1}{c|}{}                     & WTH    & \textbf{3.799}  & 1.345  & 4.920   & 1.481          & 4.221      & \textbf{1.338}       & 5.003          & 1.438                                                    \\
\multicolumn{1}{c|}{}                       & \multicolumn{1}{c|}{}                     & ECL    & \textbf{3.913}  & \textbf{1.357}  & 3.984  & 1.411          & 4.257      & 1.366      & 5.296          & 1.479                                                            \\ 
\bottomrule
\end{tabular}
}
\end{table*}

\subsubsection{Implementation details} 
\label{sssec:5-1-2}

Our experiments focus on multivariate time series forecasting task. All datasets are splited into training, validation and test set in chronological order by the ratio of 6:2:2 for ETTh1 and ETTh2 dataset and 7:1:2 for the other datasets. Our SeDAN uses 3-layer encoder and 2-layer decoder on the WTH dataset due to its large scale, and use 2-layer encoder and 1-layer decoder on the rest. The Adam is used for parameter optimization. The initial learning rate is in $\{10^{-4}, 10^{-5}\}$ and the learning rate decay and early stopping is used for 20 epochs. The hyper parameters $\lambda$ and $\gamma$ are both set to 0.1. We utilize the qhth tool \citep{64-OptNet-DBLP:conf/icml/AmosK17} to implement the differentiable convex optimization procedure in SeDAN. We use $\mathrm{MSE} = \frac{1}{n}\sum^{n}_{i=1}{(y-\hat{y})^2} $ and $\mathrm{MAE} = \frac{1}{n}\sum^{n}_{i=1}{\lvert y-\hat{y}\rvert } $ as the evaluation metrics. Our code is implemented on PyTorch and the experiments are performed on two Nvidia Titans 24GB GPUs.

\subsection{Comparison methods}
\label{ssec:5-2}

We compare our SeDAN with the following single-domain and cross-domain baselines.

\paragraph{Single-domain baselines}
We select 4 single-domain models trained only on the target dataset, including LSTNet \citep{32-RNN-LSTNet-DBLP:conf/sigir/LaiCYL18}, TCN \citep{13-TCN-DBLP:journals/corr/abs-1803-01271}, and two attention-based models: LSTMa \citep{65-LSTMa-DBLP:journals/corr/BahdanauCB14} and Informer \citep{15-informer-DBLP:conf/aaai/ZhouZPZLXZ21}.

\paragraph{Cross-domain baselines}
In the cross-domain experiments, we use the vanilla Transformer \citep{14-transformer-DBLP:conf/nips/VaswaniSPUJGKP17} model as the basic model, using the Local Time Stamp and Global Time Stamp as the position embeddings. The cross-domain baselines include the following: 1) Fine-tuning: the basic model which trains only on the source domain, and finetunes the encoder and decoder on the target domain; 2) DAN \citep{66-DAN-DBLP:conf/icml/LongC0J15}: the basic model which uses MK-MMD for adaptation on encoder features of different domain datasets without decomposition; 3) JAN \citep{62-JAN-DBLP:conf/icml/LongZ0J17}: consistent with the settings of 2), except using JMMD for cross-domain adaptation instead of MK-MMD.


\subsection{Main results}
\label{ssec:5-3}

Table~\ref{table:table-1} shows the comparative experimental results on single-domain setting, while Table~\ref{table:table-2} shows the results on cross-domain setting. In Table~\ref{table:table-1}, we can find that the proposed SeDAN outperforms all single-domain methods compared, achieving an average reduction of 27.0\% (MSE) and 16.5\% (MAE). In Table~\ref{table:table-2}, we can find that the transfer-based cross-domain baselines are usually better than the single-domain methods, which proves that the cross-domain alignment of features can help the model learn more knowledge and alleviate the problem of insufficient data. Our proposed SeDAN still achieves the best performance in cross-domain experiments. In each experiment setting, we compare our SeDAN with the best performing cross-domain baselines and obtain an average reduction of 10.9\% (MSE) and 5.3\% (MAE). From table~\ref{table:table-2}, we observe that the three transfer-based baselines do not always achieve performance improvement compared with vanilla Transformer. There may be performance decrease due to the use of source domain data, which means the negative transfer occurs. Because these cross-domain baselines adapt the marginal and conditional distributions together during the transfer process, ignoring the difficulty of transferring conditional distributions in different datasets. In contrast, our proposed SeDAN benefits from not adapting the conditional distribution within the trend features, so that the model can obtain a relatively stable performance improvement. In the case of using different source datasets, the performance of SeDAN has improved or at least remained the same, which shows its ability to effectively reduce the impact of negative transfer. This is beneficial for cross-domain transfer in time series forecasting.


\subsection{Ablation Study}
\label{ssec:5-4}

Based on ILI and ETTh2 as target datasets, we perform additional ablation experiments to demonstrate the effectiveness of each module in SeDAN.

\begin{table}[]
\caption{Ablation study of the Decomposition Adaptation method.}
\label{table:table-3}
\scalebox{0.75}{
\begin{tabular}{c|ccccccc}
\toprule
\multirow{2}{*}{Methods}     & Dataset & \multicolumn{3}{c}{ETTh2(24)}                    & \multicolumn{3}{c}{ILI(24)}                      \\ \cmidrule(lr){2-2}\cmidrule(lr){3-5}\cmidrule(lr){6-8}
                             & Source  & ETTh1          & WTH            & ECL            & ETTh1          & WTH            & ECL            \\ \midrule
\multirow{2}{*}{baseline}    & MSE     & \multicolumn{3}{c}{0.680}                        & \multicolumn{3}{c}{4.716}                        \\
                             & MAE     & \multicolumn{3}{c}{0.658}                        & \multicolumn{3}{c}{1.511}                        \\ \midrule
\multirow{2}{*}{All-JMMD}    & MSE     & 0.548          & 0.592          & 0.476          & 3.753          & 3.970          & 3.577          \\
                             & MAE     & 0.572          & 0.580           & 0.541          & 1.352          & 1.333          & 1.313          \\ \midrule
\multirow{2}{*}{All-OLA}     & MSE     & 0.412          & 0.481          & 0.443          & 3.593          & 3.426          & 3.433          \\
                             & MAE     & 0.485          & 0.544          & \textbf{0.497} & 1.231          & 1.301          & 1.278          \\ \midrule
\multirow{2}{*}{Contra2Ours} & MSE     & 0.582          & 0.662          & 0.889          & 4.049          & 3.971          & 3.691          \\
                             & MAE     & 0.609          & 0.644          & 0.746          & 1.355          & 1.347          & \textbf{1.237} \\ \midrule
\multirow{2}{*}{Ours}        & MSE     & \textbf{0.389} & \textbf{0.461} & \textbf{0.426} & \textbf{3.196} & \textbf{3.480} & \textbf{3.330} \\
                             & MAE     & \textbf{0.480} & \textbf{0.529} & \textbf{0.497} & \textbf{1.230} & \textbf{1.287} & 1.286          \\ \bottomrule
\end{tabular}
}
\end{table}

\begin{table}[]
\caption{Ablation study of the Implicit Contrastive Decomposition.}
\label{table:table-4}
\scalebox{0.75}{
\begin{tabular}{c|ccccccc}
\toprule
\multirow{2}{*}{Methods}    & Dataset & \multicolumn{3}{c}{ETTh2(24)}                    & \multicolumn{3}{c}{ILI(24)}                      \\ \cmidrule(lr){2-2}\cmidrule(lr){3-5}\cmidrule(lr){6-8}
                            & Source  & ETTh1          & WTH            & ECL            & ETTh1          & WTH            & ECL            \\ \midrule
\multirow{2}{*}{baseline}   & MSE     & \multicolumn{3}{c}{0.680}                        & \multicolumn{3}{c}{4.716}                        \\
                            & MAE     & \multicolumn{3}{c}{0.658}                        & \multicolumn{3}{c}{1.511}                        \\ \midrule
\multirow{2}{*}{DAN}        & MSE     & 0.517          & 0.467          & 0.611          & 4.649          & 3.621          & 4.172          \\
                            & MAE     & 0.571          & 0.548          & 0.640          & 1.435          & 1.277          & 1.399          \\ \midrule
\multirow{2}{*}{JAN}        & MSE     & 0.538          & 0.510          & 0.588          & 3.597          & 4.198          & 3.675          \\
                            & MAE     & 0.568          & 0.550          & 0.597          & 1.214          & 1.427          & 1.245          \\ \midrule
\multirow{2}{*}{Conv+Minus} & MSE     & 0.453          & 0.466          & 0.525          & 3.274          & 3.981          & 3.452          \\
                            & MAE     & 0.530          & \textbf{0.508} & 0.539          & 1.308          & 1.427          & 1.336          \\ \midrule
\multirow{2}{*}{Ours}       & MSE     & \textbf{0.389} & \textbf{0.461} & \textbf{0.426} & \textbf{3.196} & \textbf{3.480} & \textbf{3.330} \\
                            & MAE     & \textbf{0.480} & 0.529          & \textbf{0.497} & \textbf{1.230} & \textbf{1.287} & \textbf{1.286} \\ \bottomrule
\end{tabular}
}
\end{table}

\subsubsection{The performance of Decomposition Adaptation}
\label{sssec:5-4-1}

We first conduct ablation experiments to explore the effects of different adaptation methods to verify our hypotheses about trend and seasonal features. In experiments, we design the following variants: 1) baseline: the vanilla Transformer model with our proposed decomposition and reconstruction, but without cross-domain feature adaptation; 2) All-JMMD: the same structure of SeDAN model except that both seasonal and trend features uses joint distribution adaptation; 3) All-OLA: the same structure of SeDAN model except that both seasonal and trend features uses Optimal Local Adaptation; 4) Contra2Ours: same structure but use OLA for seasonal features, and JMMD for trend features, which is contrary to our SeDAN.

Table~\ref{table:table-3} shows the results of ablation experiments, demonstrating that SeDAN outperforms its variants in all target datasets. We observe that All-OLA performs better in all variants, second only to SeDAN. For seasonal features, compared to JMMD, OLA only transfer the marginal distribution of the features, and may lose some transferable knowledge, resulting a slight drop on performance. However, for trend features, JMMD significantly degrades the performance compared to OLA, and the transfer effect on different source datasets become unstable. The JMMD takes into account the conditional distribution while the stochastic transition dynamics are different to transfer in different domains, so that using JMMD for trend features adaptation is more likely to lead to the negative transfer.

\subsubsection{The performance of the Implicit Contrastive Decomposition.}
\label{sssec:5-4-2}

In this subsection, we verify the effectiveness of the proposed ICD. In addition to the baseline proposed in the previous subsection~\ref{sssec:5-4-1}, we design the following variants: 1) DAN or JAN: the vanilla Transformer model consistent with the description in section~\ref{ssec:5-2}, but we use DAN or JAN for undecomposed encoder features adaptation, respectively; 2) Conv+Minus: use the additive-model-based explicit sequence decomposition method, where the trend feature is directly obtained by convolution as same as SeDAN but without our contrastive learning as supervision, while the seasonal feature is obtained by subtracting the trend term from the original features.

As shown in Table~\ref{table:table-4}, our designed decomposition method outperforms Conv+Minus, proving that the proposed ICD can generate low-coupling features with more transferability. Besides, the both decomposition adaptation methods are better than DAN and JAN, which directly adapt the undecomposed features. It shows that adapting the trend and the seasonal feature separately enables the model to better learn the transferable knowledge across domains.


\begin{figure}[htbp]
  \centering
  \begin{subfigure}{.2\textwidth} 
    \centering 
    \includegraphics[width=\textwidth]{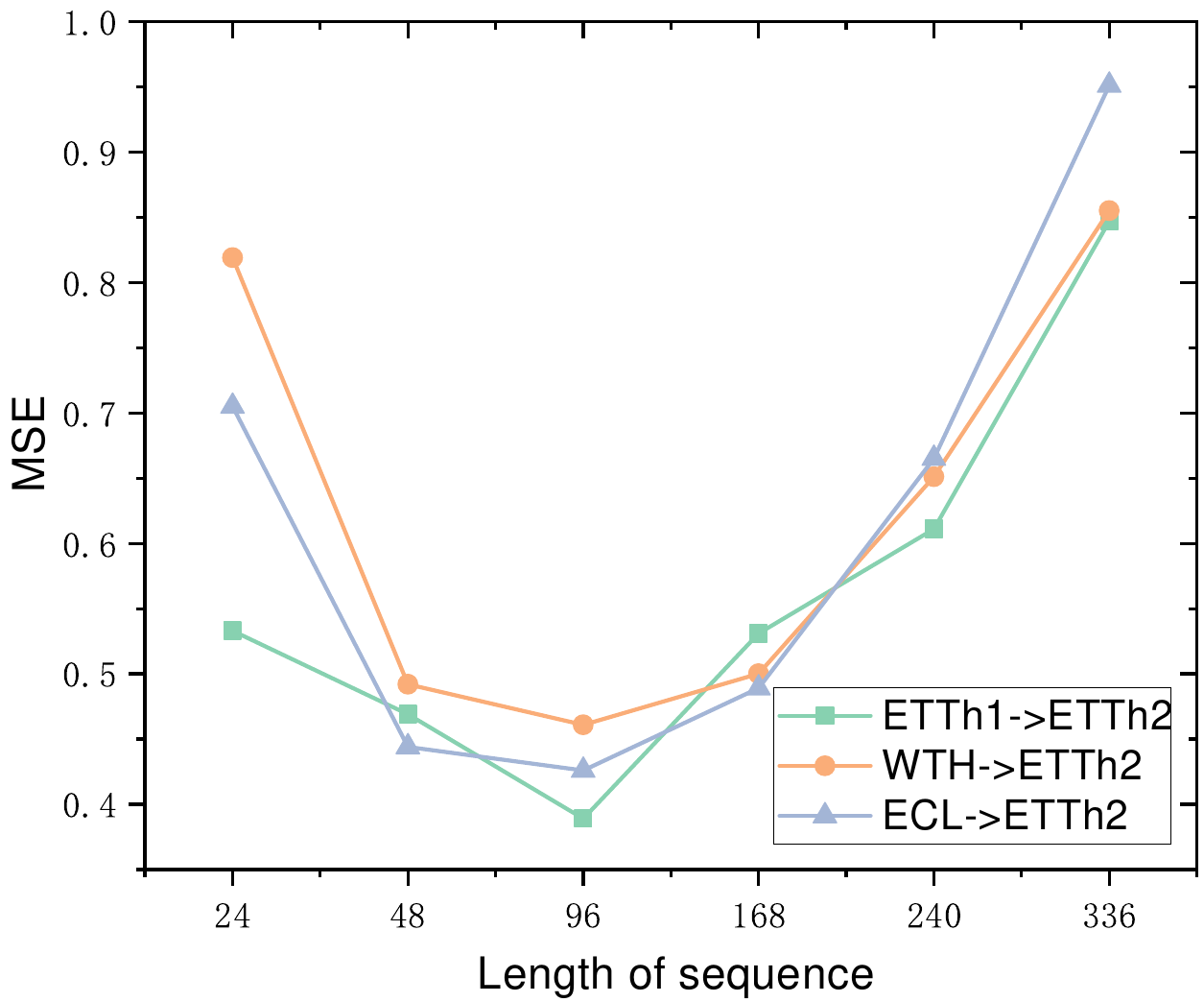} 
    \caption{predict-24 task (ETTh2).}
  \end{subfigure} 
  \begin{subfigure}{.2\textwidth} 
   \centering 
   \includegraphics[width=\textwidth]{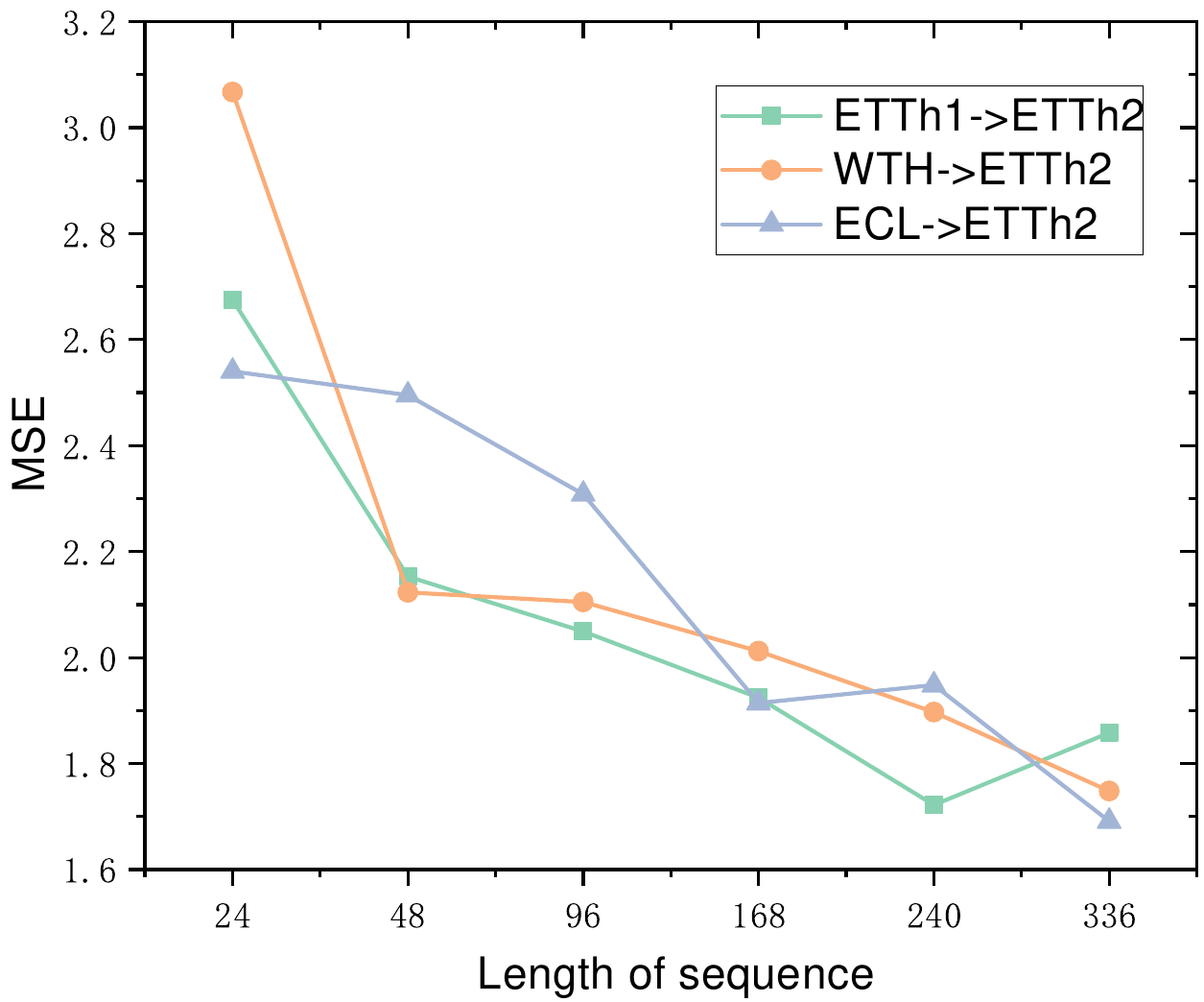} \par
   \caption{predict-168 task (ETTh2).} 
  \end{subfigure}
  \begin{subfigure}{.2\textwidth} 
   \centering 
   \includegraphics[width=\textwidth]{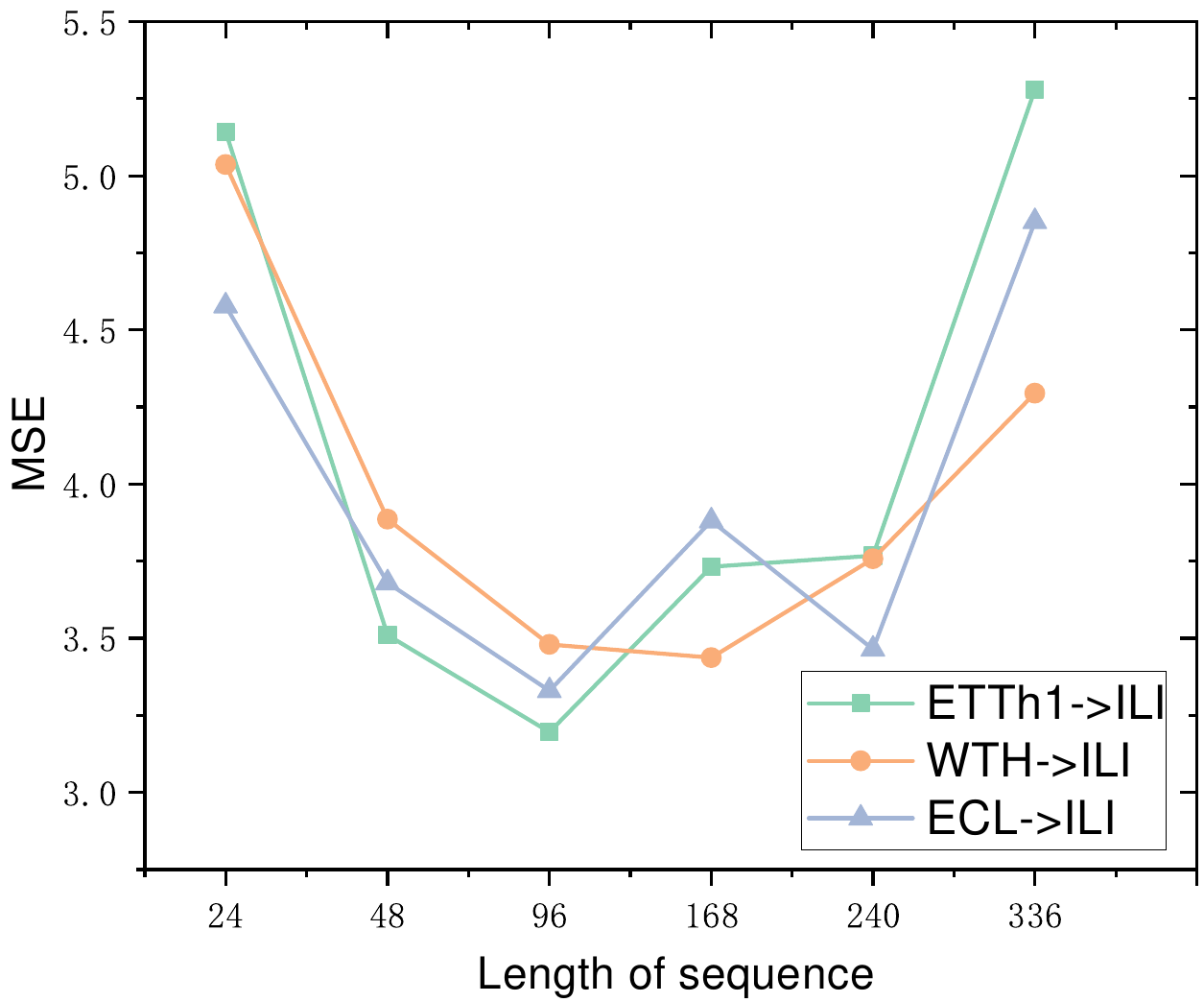}
   \caption{predict-24 task (ILI).} 
  \end{subfigure}
  \begin{subfigure}{.2\textwidth} 
   \centering 
   \includegraphics[width=\textwidth]{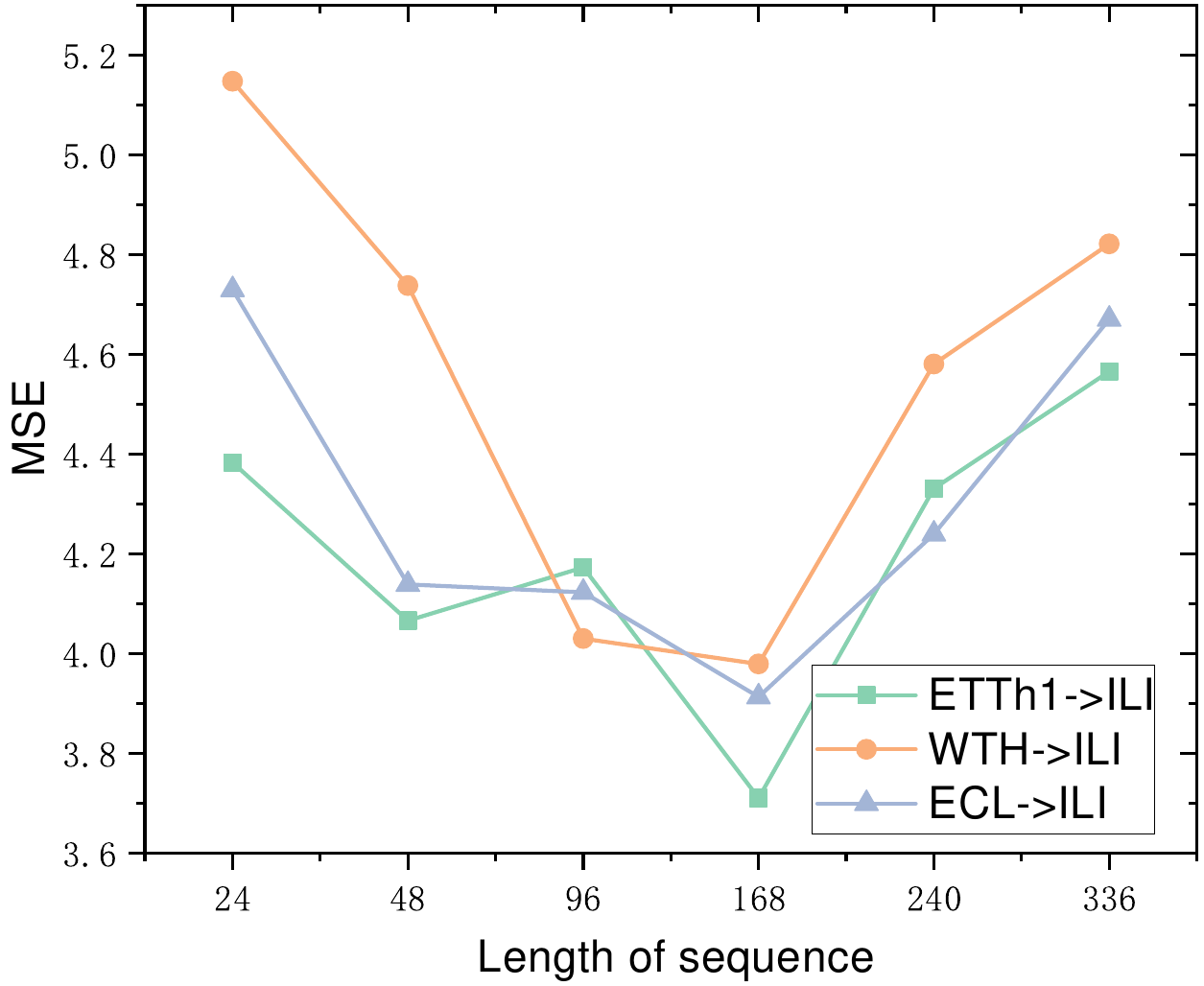} \par
   \caption{predict-60 task (ILI).} 
  \end{subfigure}
  \caption{Impact of input sequence length in SeDAN, performance is reported on ETTh2 and ILI.} 
  \label{fig:fig-5}
\end{figure}

\subsubsection{The impact of Input Sequence Length}
\label{sssec:5-4-3}

Figure~\ref{fig:fig-5} shows the impact of different input sequence lengths. At a small prediction length (like 24), with the increasing input length, the forecasting performance on ILI and ETTh2 shows the same trend: MSE first decreased, but with the further increase of input length, MSE begins to rise. Because in the beginning, the richness of the longer input information leads to an increase in the number of learnable local patterns. However, when the input length reaches a certain threshold, 
it is difficult for self-attention to query effective relevant information from complex local patterns. And the negative impact exceeds the benefit brought by the enrichment of local patterns. When predicting longer sequences, the performance of predict-60 task in ILI keeps the same trend as the predict-24 task, and its optimal input length as 168 in the predict-60 task is longer compared with 96 in the predict-24 task. However, in the predict-168 task of ETTh2, with the increase of input length, the MSE always maintains a downward trend. We consider that the reason is that for ILI, which is a small dataset recorded weekly, providing more old information will increase the difficulty of pattern query, while long-range prediction of ETTh2 requires richer local patterns.

\begin{figure}[htbp]
  \centering
  \begin{subfigure}{.2\textwidth} 
   \centering 
   \includegraphics[width=\textwidth]{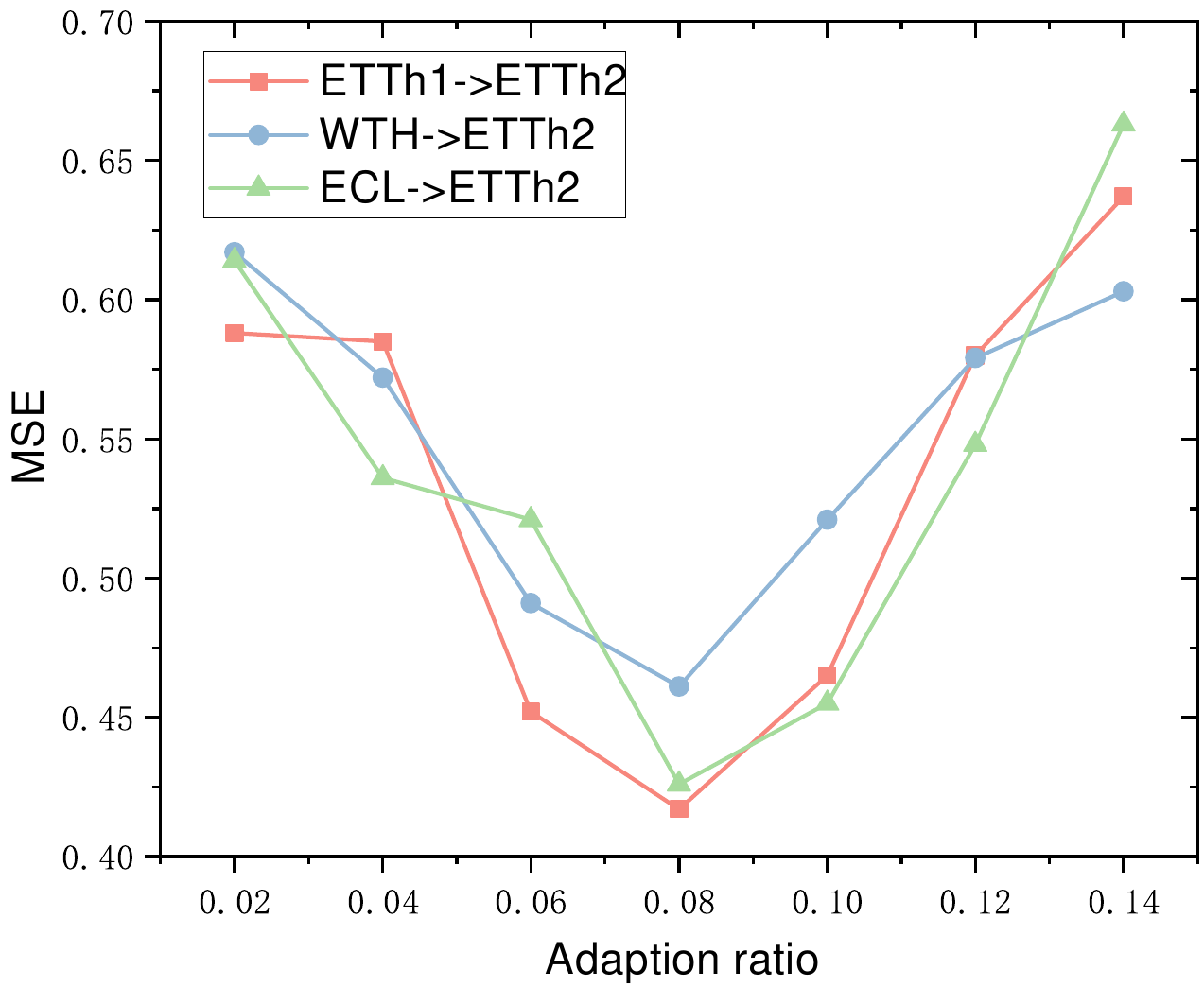}
   \caption{predict-24 task (ETTh2).} 
  \end{subfigure}
  \begin{subfigure}{.2\textwidth} 
   \centering 
   \includegraphics[width=\textwidth]{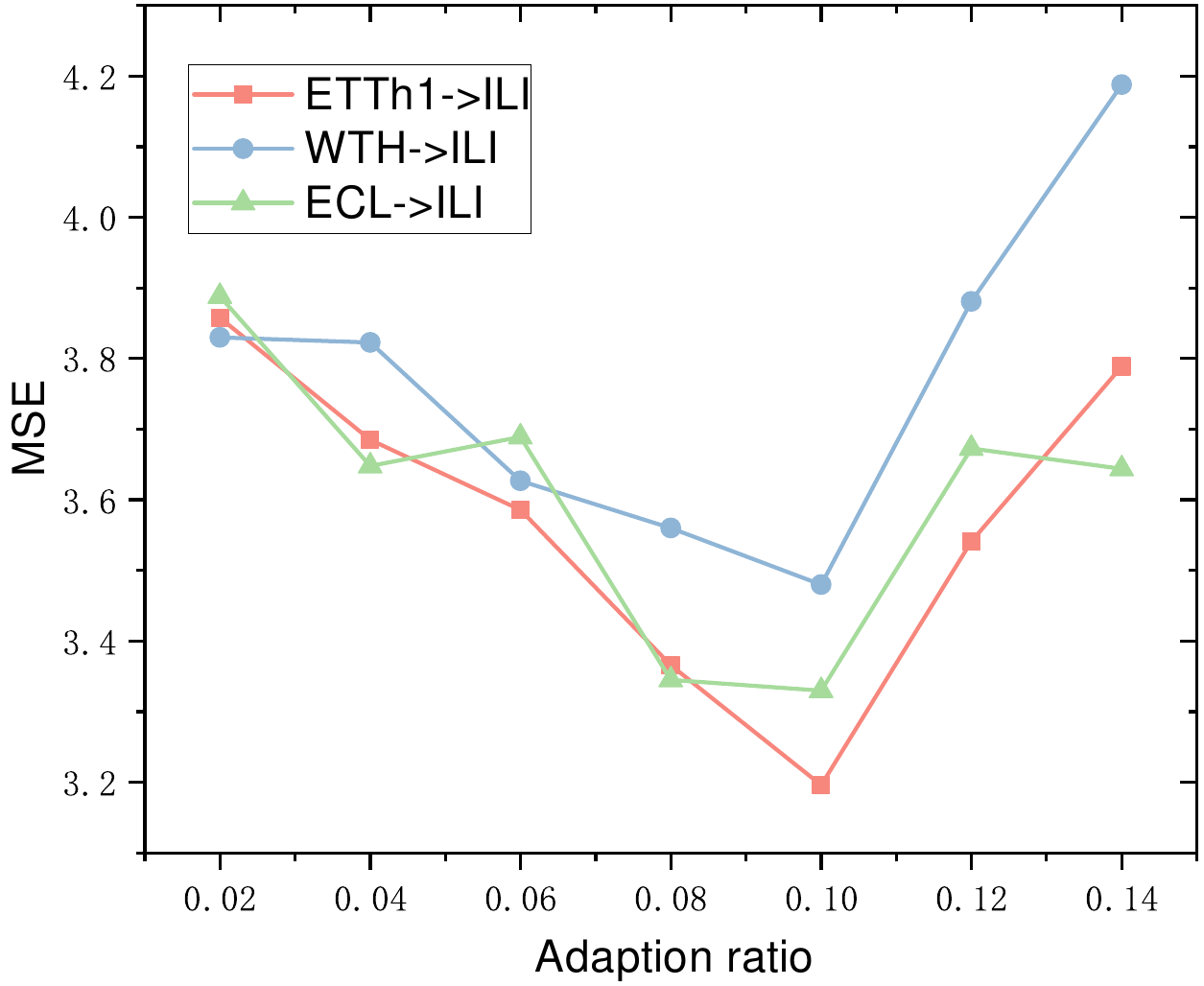}
   \caption{predict-24 task (ILI).} 
  \end{subfigure}
    \caption{Impact of loss adaption ratio in SeDAN, performance is reported on ETTh2 and ILI.} 
  \label{fig:fig-6}
\end{figure}

\subsubsection{The impact of loss adaption ratio}
\label{sssec:5-4-4}

Figure~\ref{fig:fig-6} shows the forecasting performance of the SeDAN under different adaptation ratios $\gamma$. We observe that on both datasets, as $\gamma$ increases, MSE first decreases and then increases. For ETTh2, the model achieves the best performance at $\gamma =0.08$; and for ILI, the model achieves the best performance at $\gamma =0.10$. Because the ILI dataset is smaller than ETTh2, using a larger $\gamma$ to transfer more knowledge from source domain has a positive impact on the target domain. Besides, the prediction task of ILI is more sensitive to the value of $\gamma$, and the impact of $\gamma$ on the prediction task of ETTh2 is smaller. 
Compared with ILI, there is a smaller difference between the size of ETTh2 and its source domain datasets, resulting in their different sensitivities to the value of $\gamma$.

\begin{figure}[h]
  \centering
  \begin{subfigure}{.2\textwidth} 
    \centering 
    \includegraphics[width=\textwidth]{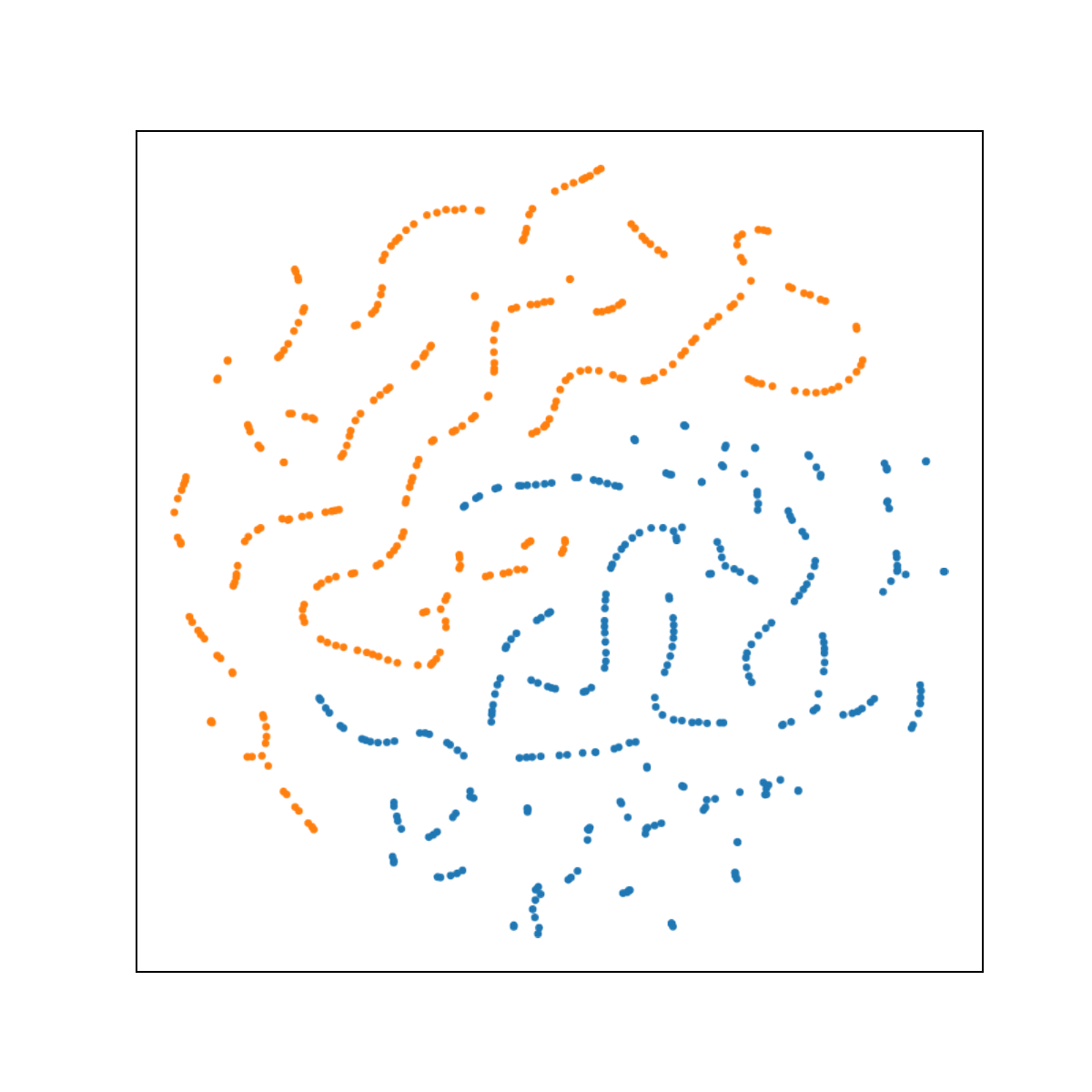} \par
    \caption{Seasonal features (before).}
    \label{fig:fig-7a}
  \end{subfigure} 
  \begin{subfigure}{.2\textwidth} 
   \centering 
    \includegraphics[width=\textwidth]{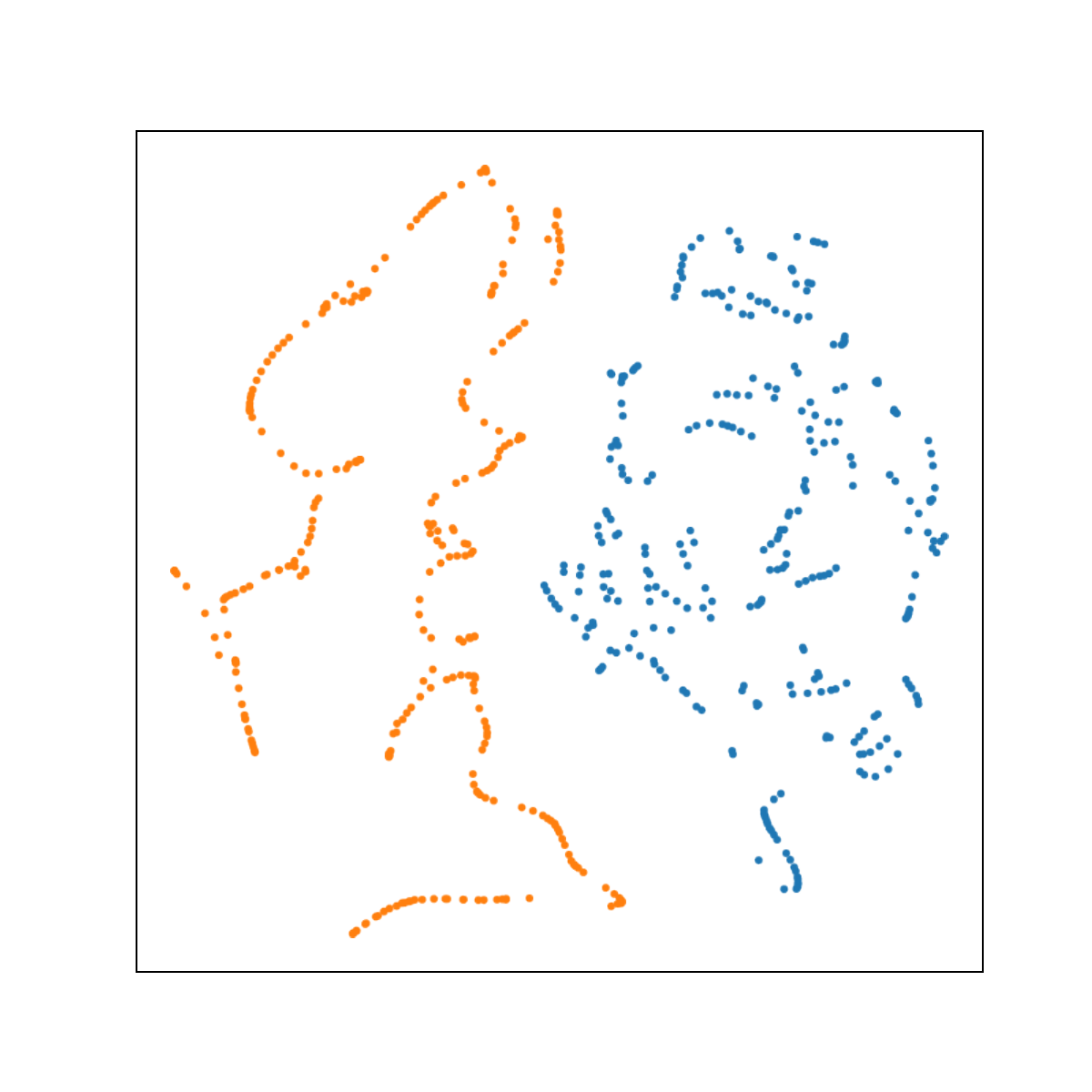} \par
    \caption{Trend features (before).} 
    \label{fig:fig-7b}
  \end{subfigure}
  \begin{subfigure}{.2\textwidth} 
   \centering 
    \includegraphics[width=\textwidth]{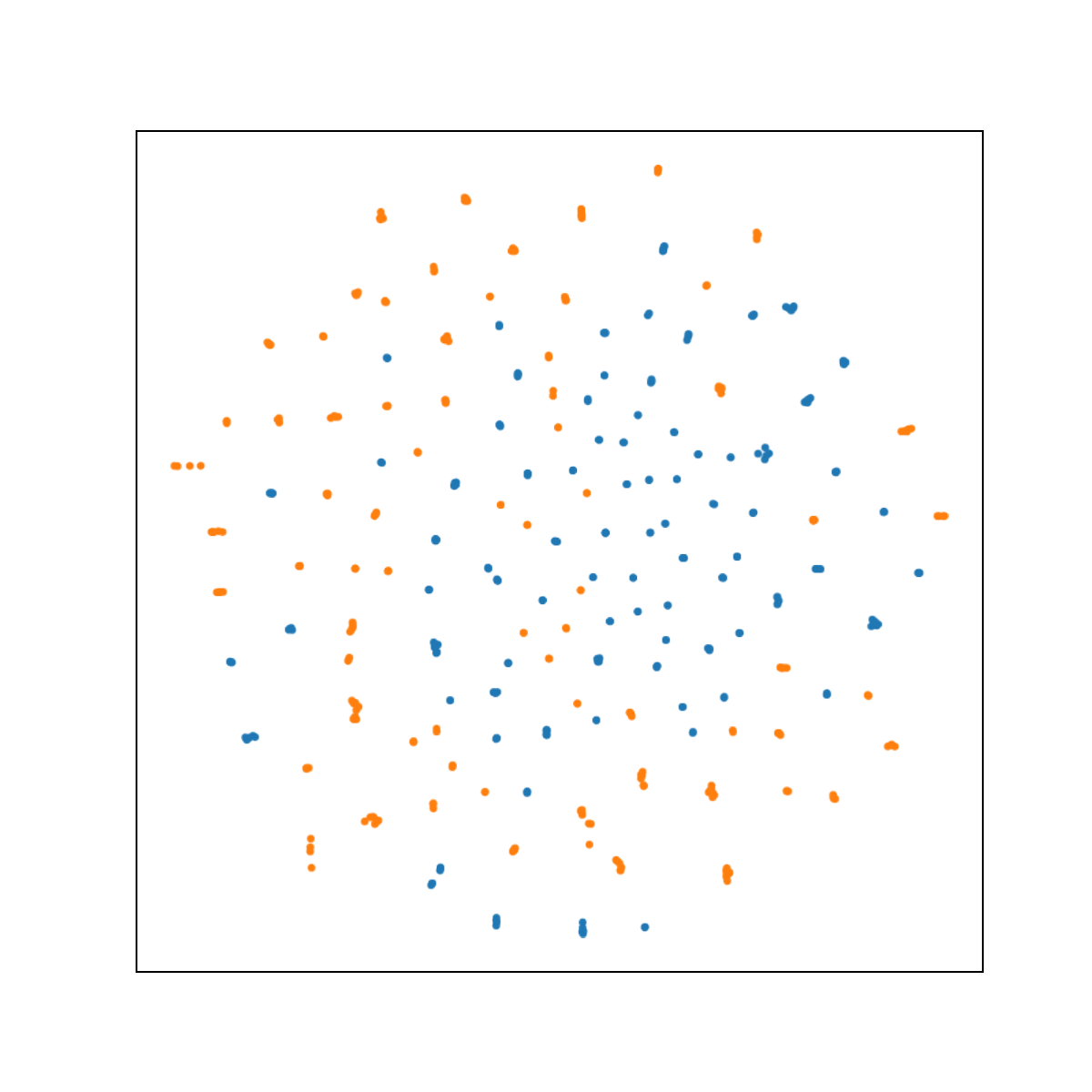} \par
    \caption{Seasonal features (after).} 
    \label{fig:fig-7c}
  \end{subfigure}
  \begin{subfigure}{.2\textwidth} 
   \centering 
    \includegraphics[width=\textwidth]{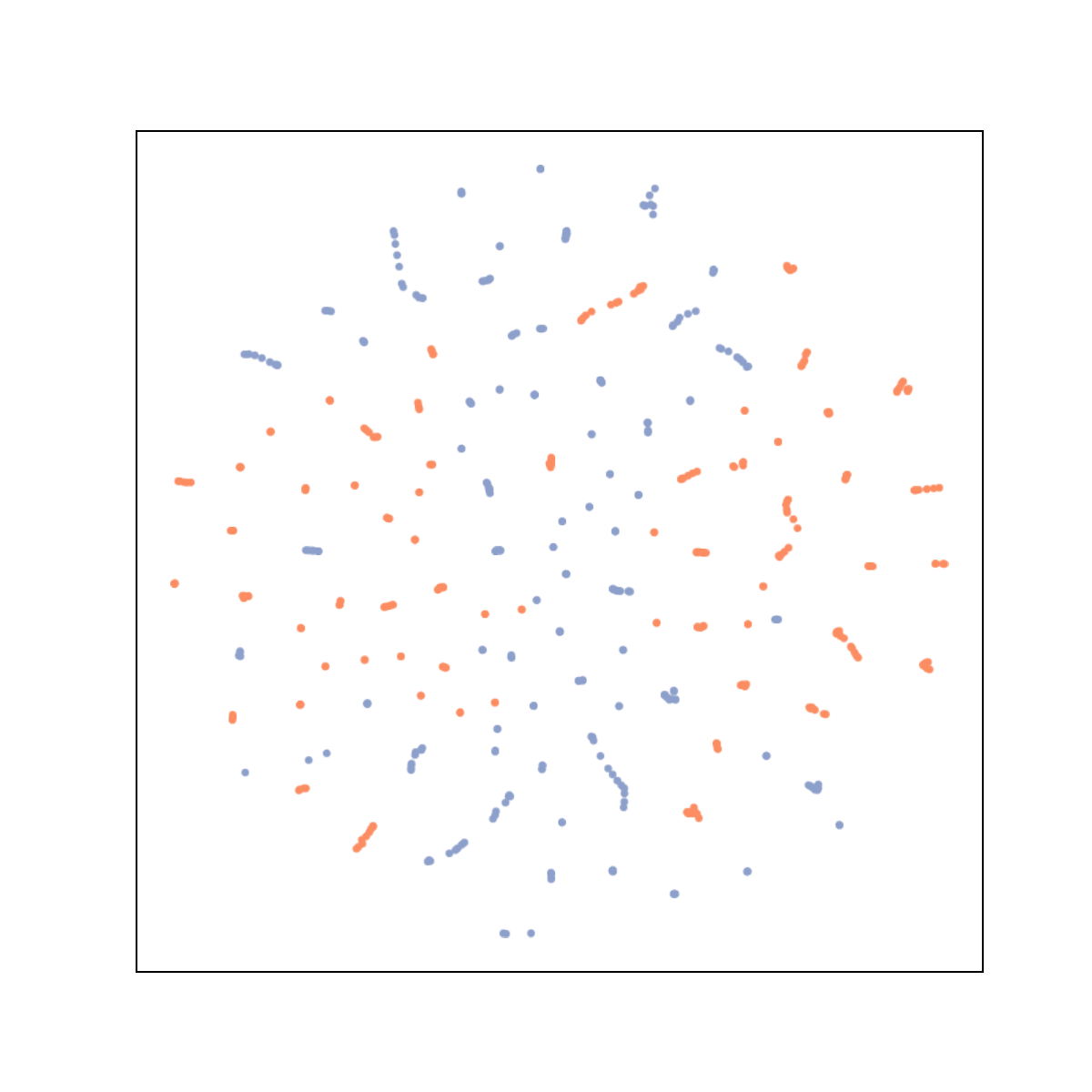} \par
    \caption{Trend features (after).} 
    \label{fig:fig-7d}
  \end{subfigure}
  \caption{Visualization of features distribution before and after decomposition adaptation. ETTh2 (orange points) as target domain and ECL (blue points) as source domain.} 
  \label{fig:fig-7}
\end{figure}

\subsection{Further Analyses on Our Method}
\label{ssec:5-5}

We use the visualization to verify the effectiveness of our SeDAN, use the t-SNE to analyze the influence of proposed Decomposition Adaptation on the feature distribution and further discuss the transferability between datasets.

\paragraph{Adaptation Analysis in Feature Space}
We visualize the feature space adaptation on the ETTh2 dataset to illustrate the effectiveness of the Decomposition Adaptation method. Figure~\ref{fig:fig-7a} and Figure~\ref{fig:fig-7b} show the distribution of seasonal and trend features in the source and target domain when only decomposition is performed without adaptation. By comparison, it can be found that the trend features have a certain degree of continuity in the feature space compared with the seasonal features. As shown in Figure~\ref{fig:fig-7c} and Figure~\ref{fig:fig-7d}, after feature decomposition in SeDAN, the corresponding features are better adapted by proposed appropriate adaptation methods. Moreover, the distance between the cross-domain features after reconstruction is effectively shortened. The negative transfer problem caused by the non-transferable components in the features is avoided, so that the overall transfer performance is better and more stable.

\paragraph{When to transfer}

This paper mainly focuses on what type of knowledge can be transferred across domains in time series forecasting, and how to transfer these knowledge. In this subsection, we briefly analyze the problem of ``When to transfer", which means the performance impact of different sources on the same target domain. As shown in Table~\ref{table:table-2}, when ETTh1 and ETTh2 are chosen as source domains of each other, the performance is better than that using other datasets as source domains because their fields, sampling methods and scales are all similar. Furthermore, the transfer performance tends to be better when the size of the source dataset is larger, possibly due to the increase in the transferable knowledge and local patterns that can be matched. Therefore, we consider that starting from the similarity of the local patterns contained in the dataset is a way to deal with ``When to transfer".

\section{Conclusion}
\label{conclusion}

In this paper, we explore the transferability of knowledge in cross-domain time series datasets. Starting from the basic problems of transfer learning, we argue the transferable knowledge types between different time series domains, and propose a general transfer framework SeDAN for multivariate time series forecasting. 
Through the feature-level ICD, we use contrastive learning to decompose the seasonal and trend features. According to their respective transferability, we propose the corresponding decomposition adaptation methods: JMMD for the seasonal features and OLA for the trend features. 
We conduct experiments on real-world datasets and demonstrate that the proposed SeDAN model can improve the prediction performance on the target domain with the use of different source domains. Compared with general transfer learning methods, SeDAN can provide a more stable transfer effect. 
This paper mainly focuses on the transferable knowledge types and corresponding transfer methods. In the future, we plan to further explore the problem of ``When to transfer" from the perspective of time series similarity.

\bibliographystyle{named}
\bibliography{ijcai20}

\end{document}